\documentclass[10pt,journal,compsoc]{IEEEtran}
\usepackage{preamble}

\begin{document}


%
\title{Towards Human-centered Explainable AI: A Survey of User Studies for Model Explanations}
%
%
%
%

\author{Yao~Rong, Tobias~Leemann, Thai-Trang~Nguyen, Lisa~Fiedler, Peizhu~Qian, Vaibhav~Unhelkar, Tina~Seidel, Gjergji~Kasneci, and~Enkelejda~Kasneci
\IEEEcompsocitemizethanks{\IEEEcompsocthanksitem Y.~Rong, Tina~Seidel, Gjergji~Kasneci and Enkelejda~Kasneci
are with Technical University of Munich, 80335, Munich, Germany. E-mail: \{yao.rong, tina.seidel, gjergji.kasneci, enkelejda.kasneci\}@tum.de}
\IEEEcompsocitemizethanks{\IEEEcompsocthanksitem Tobias~Leemann, Thai-trang~Nguyen, and Lisa~Fiedler are with University of T\"ubingen, 72076, T\"ubingen, Germany.
E-mail: \{tobias.leemann\}@uni-tuebingen.de, \{thai-trang.nguyen, lisa.fiedler\}@student.uni-tuebingen.de} 
\IEEEcompsocitemizethanks{\IEEEcompsocthanksitem Peizhu~Qian and Vaibhav~Unhelkar are with Rice University, 77005, Houston, USA. E-mail: \{pq3, vaibhav.unhelkar\}@rice.edu}
\thanks{Manuscript received 29 January 2023. Corresponding author: yao.rong@tum.de}} %

%
%

\markboth{Journal of \LaTeX\ Class Files,~Vol.~14, No.~8, August~2015}%
{Shell \MakeLowercase{\textit{et al.}}: Bare Demo of IEEEtran.cls for Computer Society Journals}
%




\IEEEtitleabstractindextext{%
\begin{abstract}
Explainable AI (XAI) is widely viewed as a sine qua non for ever-expanding AI research. A better understanding of the needs of XAI users, as well as human-centered evaluations of explainable models are both a necessity and a challenge. In this paper, we explore how human-computer interaction (HCI) and AI researchers conduct user studies in XAI applications based on a systematic literature review. After identifying and thoroughly analyzing \numcorepapers core papers with human-based XAI evaluations over the past five years, we categorize them along the measured characteristics of explanatory methods, namely \revise{\textit{trust, understanding, usability}, and \textit{human-AI collaboration performance}}. Our research shows that XAI is spreading more rapidly in certain application domains, such as recommender systems than in others, but that user evaluations are still rather sparse and incorporate hardly any insights from cognitive or social sciences. Based on a comprehensive discussion of best practices, i.e., common models, design choices, and measures in user studies, we propose practical guidelines on designing and conducting user studies for XAI researchers and practitioners. Lastly, this survey also highlights several open research directions, particularly linking psychological science and human-centered XAI. 
\end{abstract}

}


\maketitle

\IEEEdisplaynontitleabstractindextext

%

\IEEEpeerreviewmaketitle

\IEEEraisesectionheading{\section{Introduction}\label{sec:introduction}}

\begin{table*}[t]
 \setlength{\belowcaptionskip}{-4mm}
\centering
\resizebox{.98\linewidth}{!}{
\begin{tabular}{rrc}
    \toprule
    \textbf{Trust} & &  \begin{tabular}[c]{@{}c@{}} \cite{panigutti2022understanding, anik2021data, colley2021effects, ehsan2021expanding, liao2021should, tsai2021exploring, guo2022building, ooge2022explaining, suresh2022intuitively, paleja2021utility, schaffer2019can, wang2021explanations, buccinca2020proxy, peng2022inherently, zhang2020effect} \\ \cite{dominguez2019effect, cai2019effects, millecamp2019explain, tsai2018beyond, li2019data, kaur2020interpreting, cheng2019explaining, kunkel2019let, kim2020answering, lai2019human, rong2022user, schoeffer2022there, ehsan2019automated, smith2020no, smith2020digging, springer2019progressive} \end{tabular} \\
    \midrule
    \multirow{3}{*}{\textbf{Understanding}} & subjective & \cite{ross2021evaluating, radensky2022exploring, hadash2022improving, chromik2021think, cheng2019explaining, rebanal2021xalgo, kuhl2022keep, wang2021explanations, buccinca2020proxy, cai2019effects, ehsan2019automated, dominguez2019effect, rader2018explanations, bell2022accuracyexplain, guo2022building, hase2020evaluating, schuff2022human, bang2021explaining, kim2022hive, peng2022inherently,szymanski2021visual,springer2019progressive} \\
    & objective & \cite{plumb2020regularizing, hase2020evaluating, zhang2022towards, ross2021evaluating, bove2022contextualization, abdul2020cogam, ramamurthy2020model, arora2022explain, antoran2021getting, borowski2021exemplary, buccinca2020proxy, wang2021explanations, poursabzi2021manipulating, alqaraawi2020evaluating, ribeiro2018anchors, nourani2021anchoring, chromik2021think, bell2022accuracyexplain, sixt2022do, cheng2019explaining, chandrasekaran2018explanations, colin2022what, kim2022hive, shen2020useful,szymanski2021visual}\\
    & explanation model & \cite{yeh2019completeness, ghorbani2019towards, leemann2022coherence, laina2020quantifying, zhang2022towards, kaur2020interpreting, wang2022interpretableideation, ramamurthy2020model}\\
    \midrule
    \multirow{5}{*}{\textbf{Usability}}
    & workload & \cite{abdul2020cogam, kaur2020interpreting, dominguez2019effect, colley2021effects,arendt2020parallel} \\
    & helpfulness & \cite{zhang2022towards, abdul2020cogam, nourani2021anchoring, wang2022interpretableideation, zhang2022debiased, buccinca2020proxy, gao2019explainable, plumb2020regularizing}\\
    & satisfaction & \cite{bove2022contextualization, dominguez2019effect, guo2022building, millecamp2019explain, tsai2021exploring, kouki2019personalized, panigutti2022understanding, smith2020no, tsai2018beyond, tsai2019explaining} \\
    & \revise{undesired behavior detection} & \cite{poursabzi2021manipulating, kim2020answering, balayn2022can, sixt2022do, rawal2020beyond, rader2018explanations, grgic2018human, dodge2019explaining, harrison2020empirical, schoeffer2022there, wang2022humans, anik2021data, htun2021perception, binns2018s, schoeffer2021appropriate} \\
    & ease of use and others & \cite{ross2021evaluating, balayn2022can, kaur2020interpreting, buccinca2020proxy, abdul2020cogam, donkers2020explaining, kuhl2022keep, hohman2019gamut, smith2020digging, wang2022interpretableideation, kim2020answering, colley2021effects, kuhl2022let, schneider2021explain, panigutti2022understanding,arendt2020parallel,choi2018fontmatcher,le2018improving, li2019data,shang2022not,dodge2022people}\\
    \midrule
    \revise{\textbf{Human-AI Collaboration Performance}} & & \cite{das2020leveraging, paleja2021utility, bansal2021does, lai2019human, lai2020chicago, feng2019can, alufaisan2021does, buccinca2020proxy, zhang2020effect, gajos2022people, nourani2021anchoring, smith2020digging, liao2022user, bell2022accuracyexplain, lai2019human, poursabzi2021manipulating, smith2020no, nguyen2021the, taesiri2022visual, kim2022hive, nguyen2022visual} \\
    \bottomrule
\end{tabular}
}
\caption{Overview of the core papers containing user studies in XAI grouped by categories of measurements. \rev{As some core papers assess quantities belonging to several groups, a single paper can also be listed among multiple groups.}
\label{tab:categories}}
\end{table*}

Artificial Intelligence (AI) is driving digital transformation and is already an integral part of various everyday technologies. Recent developments in AI are essential to progress in fields such as recommendation systems \cite{wei2017collaborative,yang2017combining,zhang2018towards}, 
autonomous driving \cite{grigorescu2020survey,cui2019multimodal,rong2021artificial} 
or robotics \cite{murphy2019introduction,rajan2017towards,wachter2017transparent}. Moreover, AI's success story has not excluded high-stakes decision-making tasks like medical diagnosis \cite{park2018methodologic,sidey2019machine,vaishya2020artificial}, 
credit scoring \cite{dastile2020statistical,ala2022deep,addo2018credit}, 
jurisprudence \cite{van2019crowdsourcing,sourdin2018judge} 
or recruiting and hiring decisions \cite{raghavan2020mitigating,tambe2019artificial}, 
However, the behavior and decision-making processes of modern AI systems are often not understandable, so they are frequently considered black boxes. 
\revise{Deploying such black-box models presents a serious dilemma in certain safety-critical domains, for instance, public health or finance~\cite{castelvecchi2016can}. This is due to the necessity for a transparent and trustworthy AI system, which is required by both practitioners (to gain better insights into system functioning) and end users (to rely on model decisions)}.
 

Methods to increase the interpretability and transparency of an AI system are developed in the research area of Explainable AI (XAI). Specifically, human-centered XAI, which addresses the importance of human stack-holders to the AI systems, has been proposed and discussed since \cite{riedl2019human, ehsan2020human}. While a huge number of model explanations are available, 
the question of how to transparently evaluate their quality is still an open research question, and hence, extensively studied in recent years.
A popular taxonomy of evaluation strategies for XAI methods proposes three categories: functionally-grounded evaluation, application-grounded evaluation, and human-grounded evaluation \cite{doshi2017towards}. 
While functionally-grounded measures 
do not require human labor, the other two involve human subjects and are more costly to conduct.

Many functionally-grounded measures have been proposed to evaluate XAI algorithms (see \cite{nauta2023anecdotal} for review), however, the difficult comparability between different automatic evaluation measures is a common problem \cite{tomsett2020sanity,rong2022consistent}. Another drawback of automated measures is that there is no guarantee that they truly reflect humans' preferences \cite{hase2020evaluating,nguyen2018comparing}. Consequently, user studies in XAI, especially when moving towards real-world products, are inevitable if one wishes to test more general beliefs of the quality of explanations \cite{dominguez2019effect}. However, only a small portion (about 20\,\%) of XAI evaluation projects consider human subjects \cite{nauta2023anecdotal}. There exist efforts in developing taxonomies or introducing the definitions or implications of different human-centric evaluations \cite{hoffman2019evaluating, chromik2020taxonomy, mohseni2021multidisciplinary}, but the recent generation of user studies and their findings have not been systematically discussed yet. Moreover, Yang et al.~\cite{yang2018mapping} point out that XAI is growing separately and treated differently in different communities (e.g., machine learning and HCI). Hence, effective guidance in XAI user study design is crucial to better let both XAI algorithm and application designers recognize the users' real needs. This work aims to bridge this research gap in modern XAI user study design by distilling practical guidelines for user studies through a comprehensive and structured literature review.

\rev{Therefore, we reviewed highly relevant papers that include user studies from top-tier HCI and XAI venues. Specifically, we included the recent \textit{five} years of CHI, IUI, UIST, CSCW, FA(cc)T, ICML, ICRL, NeurIPS, and AAAI. 
As we aim at analyzing human user evaluation of advanced model explanations, we ran search queries involving keywords from the two groups ``explainable AI'' and ``user study'', as listed in the \Cref{tab:keywords}. We selected the papers containing at least one keyword from each group, resulting in over one hundred papers. Then, we thoroughly studied these papers and filtered out papers that did not fulfill the criteria: (1) deploying explainable models or techniques and (2) conducting an assessment with human subjects.}
We identified a total of \numcorepapers core papers for this survey (see \Cref{tab:categories} for an overview of core papers with respect to their measured quantities in user studies). Based on these core papers, we performed a comprehensive analysis to fill the research gap by offering a systematic overview of user studies in XAI. We highlight the main contributions:
\begin{enumerate}
    \item To offer an overview of the foundational work of user studies in XAI, we investigated references of all \numcorepapers core papers in a data-driven manner. Likewise, we analyzed follow-up works building on these core papers (identified through citations of core papers) to reveal the fields impacted by XAI user evaluations (\Cref{sec:part1}). 
    \item We present a summary of the design details in XAI user studies with particular focus on the deployed models and explanation techniques, experimental design patterns, participants as well as concrete measures, providing inspiration of how to collect human assessment (\Cref{sec:part2}).
    \item We discuss the impact of using explanations on different aspects of user experience (\Cref{sec:previous findings}), which can serve as an overview of the effectiveness of the current XAI technology and a summary of the state-of-the-art.
    \item Based on the examined user study details and their best-practice findings, we synthesize guidelines for designing an effective user study for XAI (\Cref{sec:part3}).
    \item Beyond the user study design, we discuss potential paradigms of AI systems understanding humans in the context of e.g., theory of minds, as well as other future research directions (\Cref{sec:summary}).
\end{enumerate}

Our study highlights under-investigated areas in the context of current user-centered XAI research such as cognitive or psychological sciences through data-driven bibliometric analysis. Together with our proposed guidelines, we believe that this work will benefit XAI practitioners and researchers from various disciplines and will help to approach the overarching goal of human-centered XAI.

\begin{table}[]
\setlength{\abovecaptionskip}{0.5mm}
\setlength{\belowcaptionskip}{-6mm}
\resizebox{\linewidth}{!}{
\begin{tabular}{c|c|c}
\hline
         & \textbf{Explainable AI}    & \textbf{User Study}  \\ \hline
\textbf{Keywords} & \begin{tabular}[c]{@{}c@{}}XAI, explainable AI,  \\ explanation,  explainable, \\explanatory, interpretable, \\ intelligible,  black-box,  \\ machine learning,\\ explainability, interpretability, \\ intelligibility, explain \\ attribution, feature \end{tabular} & \begin{tabular}[c]{@{}c@{}}user study, participant , \\ human subject, \\  empirical study, \\ lab study, \\ user evaluation, \\ human evaluation \end{tabular} \\\hline
\end{tabular}
}
\caption{Keywords for our paper search query. Two groups of keywords were used.}
\label{tab:keywords}
\end{table}

\begin{figure*}[h]
\centering
 \setlength{\belowcaptionskip}{-4mm}
\includegraphics[width=.95\linewidth]{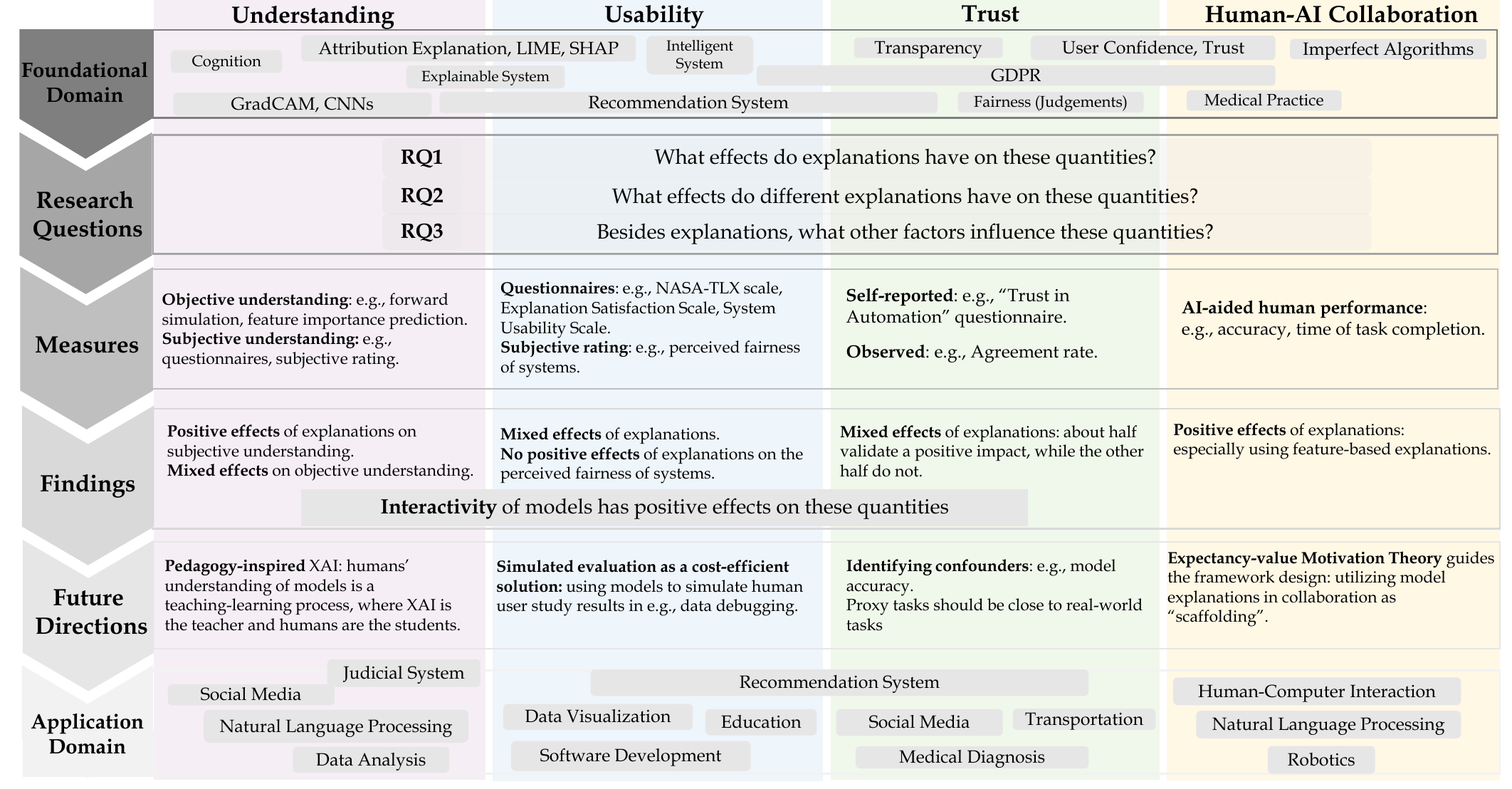}
\caption{\rev{Roadmap of our literature analysis. We find out the foundational works of core papers and their application domains using a data-driven method introduced in~\Cref{sec:part1}. Three main research questions in user studies are distilled from core papers. Methods related to measures of each category are discussed in~\Cref{sec:part2}, and findings of the research questions are summarized in~\Cref{sec:previous findings}. Based on the findings, we propose future directions to further promote human-centered XAI in~\Cref{sec:summary}. We distill important messages in this figure, but refer to the discussion in the corresponding sections for more details.}}
\label{fig:combined}
\end{figure*}

\vspace{-5pt}
\section{Related Work}
\label{sec:related work}
As a vast amount of explanation methods have been proposed, many researchers seek a systematic overview of the ever-growing field of XAI. In \cite{adadi2018peeking, arrieta2020explainable,samek2019towards,burkart2021survey,carvalho2019machine,gilpin2018explaining}, the authors aim to cover many facets of XAI technologies ranging from problem definitions, goals, AI/ML model explanations to evaluation measures, while in \cite{abdul2018trends} the authors emphasize the research trends and challenges in Human-Computer-Interaction (HCI) applications. A large body of XAI surveys focuses mainly on the interpretability of a particular family of models and corresponding explanation techniques. For instance, \cite{montavon2018methods,das2020opportunities,joshi2021review} investigate explanations for Deep Neural Networks (DNNs), where models often take images as input \cite{montavon2018methods,das2020opportunities}. Joshi et al.~\cite{joshi2021review}, however, provide an extensive review for DNNs with multimodal input for instance that of joint vision-language tasks. Causal interpretable models are gaining more attention recently and Moraffah et al.~\cite{moraffah2020causal} provide a literature review for causal explanations. 
A systematic literature review on explanations for advice-giving systems is conducted in \cite{nunes2017systematic}. 
\revise{Among these surveys focusing on general XAI technologies, evaluation measures are only briefly examined.} 

One challenge in XAI research is to evaluate and compare different explanation methods, due to the multidisciplinary concepts in interpretability/explainability \cite{nauta2023anecdotal, doshi2017towards,lipton2018mythos}. 
Evaluation measures can be divided into two groups: human-grounded measures that rely on human subjects and functionally-grounded metrics that can be computed without human subjects \cite{doshi2017towards,nauta2023anecdotal}. 
Many researchers seek solutions to evaluate explanations automatically. A comprehensive literature review with a focus on these functionally-grounded evaluation methods (without human subjects) can be found in \cite{nauta2023anecdotal}. 
Explainability is an inherently human-centric property, therefore, the research community should and has started to recognize the need for human-centered evaluations when working on XAI \cite{doshi2017towards, liao2021human}.

For instance, Chromik and Schuessler~\cite{chromik2020taxonomy} propose a taxonomy on XAI evaluations involving humans. Mohseni et al.~\cite{mohseni2021multidisciplinary} summarize four groups of human-related evaluation metrics: mental model (e.g., user's understanding of the model), user trust, human-AI task performance and explanation usefulness and satisfaction (i.e., user experience). Hoffman~\cite{hoffman2019evaluating} places more focus on psychometric evaluations by proposing a conceptual model of the XAI process and specifying four key components that should be evaluated: explanation goodness and satisfaction, (user's) mental models, curiosity, trust and performance. Beyond assessing evaluation methods, XAI applications are designed to eventually support decision-making and benefit end users. 
A recent review by Lai et al.~\cite{lai2021towardshumanai} considers studies on collaborative Human-AI decision-making, which may include AI agents providing explanations. Success in human-AI decision-making tasks can be seen as one amongst many other ways to evaluate the effect of explanations. 
Ferreira and Monteiro~\cite{ferreira2020people} present a review of the user experience of XAI applications to answer who uses XAI, why, and in which context (what + when) the explanation is presented.

Closer to our focus on user studies concerning XAI, Liao et al.~\cite{liao2021human} study user experiences with XAI to reveal pitfalls of existing XAI methods, underscoring the important role of humans in XAI development. 
As suggested by Doshi-Velez and Kim~\cite{doshi2017towards}, a human-subject experiment needs to be designed sophisticatedly to reduce confounding factors. 
\revise{In contrast to previous surveys on XAI, we aim to provide XAI researchers and practitioners with a comprehensive overview of the research questions explored in user studies, along with thorough information on experimental design. To this end, we present a practical guideline in user study design, which can be used as a starting point for future exploration of human-centric XAI applications.}

\section{\rev{Methodology}}
\label{sec:part1}
\rev{To analyze the collected papers related to user studies on XAI, we first categorize them into four groups based on their objectives. From these studies, we distill three main research questions concerning the effects of model explanations on each objective. We then summarize the methods used in these studies to quantify these objectives. Important findings from the papers are discussed, and we propose future directions based on these findings. Additionally, we examine the foundational works upon which these user studies are based (i.e., their references) and the follow-up papers that cite them, shedding light on the foundational works and emerging trends in human-centered XAI studies. \Cref{fig:combined} presents a roadmap of our analysis.}

\rev{In this section, we first describe the criteria used for their categorization. We then discuss the foundational and application domains of these papers, providing a broader view before diving into their detailed analysis.}
\subsection{\rev{Categorization of User-Study Objectives}}
\label{sec:method}

Since the core papers cover various factors of model explanations, we decided to categorize the core papers into different clusters to better study their commonalities and differences. \revise{In \cite{doshi2017towards}, \textit{interpretability} in the context of ML systems is defined as the ability to explain or present model predictions in understandable terms to a human. Beyond fostering comprehension, the authors argue that interpretability can assist in qualitatively ascertaining whether other desiderata, such as \textit{usability} and \textit{trust} are met. During a profound study of the relevant literature that was previously selected, we identified four sensible categories, that are derived from the considered dependent variables in user studies (desiderata of interpretability). These four categories are \textbf{trust, understanding, usability}, and \textbf{human-AI collaboration performance}}.
In \Cref{tab:categories}, the studied papers are categorized according to the measured quantities. \rev{As each measure can usually be assigned to only one of these categories, we found this distinction to be intuitive.}

\rev{These categories reflect different functionalities (goals) of XAI. As interpretability is defined as ``\textit{the ability to explain or to present in understandable terms to a human.}'', humans' ``understanding'' is the direct goal of XAI.} To be concrete, understanding in the context of interacting with an ML model refers to a user's grasp or ``mental model'' of how the model operates, and this knowledge grows from using the system and from clear explanations about it~\cite{liao2021human}. \rev{``Usability'' is commonly studied in human-computer interaction~\cite{bevan2001international}, which is one of the desiderata of XAI~\cite{doshi2017towards}. According to~\cite{iso1998ergonomic}, usability is the extent to which users can utilize a product to successfully, efficiently, and satisfactorily accomplish their intended objectives.} Thus, this category encompasses user studies that employ model explanations to support users in achieving specific tasks.
In usability, different aspects are measured, for instance, whether the system is easy to use or how much cognitive load it requires. 
The aspect ``undesired behavior detection'' relates to use cases where explanations uncover model discriminatory behaviors, such as the utilization of undesired features.
\rev{``Trust'' in AI is summarized as a combination of the user's confidence in a model’s accuracy, a personal comfort level with understanding and using it, and the willingness to let the model make decisions~\cite{lipton2018mythos}. It encompasses more requirements.}
\revise{Human-AI collaboration} performance is related to scenarios where the AI system provides its predictions, but humans retain the final decisions \cite{bansal2021does}. In this case, model explanations are deployed to reach a performance superior to that of the AI system or the human decision-maker alone.
\revise{These categories cover different dependent variables of interest in the reviewed user studies, primarily related to how XAI methods function. These functions mainly tie to the models' reasoning and knowledge representation. A wider perspective on XAI, which assesses generalization or robustness, remains an important field for future exploration through user studies.}
\vspace{-7pt}
\subsection{Foundations of User Studies}
\label{sec:foundation}
\rev{Based on a data-driven bibliometric analysis of the references in core papers, we highlight significant research topics within the "Foundational Domain" in \Cref{fig:combined}. It is evident that model explanations and interpretability are pivotal components. This includes papers that introduce explanation methods such as LIME~\cite{ribeiro2016should}, SHAP~\cite{lundberg2017unified}, and other attribution methods. These are a frequent subject of study in works measuring understanding and usability.} Additionally, convolutional networks, which are commonly employed in experiments, use tools like GradCAM~\cite{selvaraju2017grad} and various saliency maps to generate model explanations.
Notably, many research papers appear within the domain of recommender systems, because many XAI user studies are conducted in the context of recommendation solutions. 
The EU's General Data Protection Regulation (GDPR) \cite{voigt2017eu} is frequently mentioned in core papers due to the ongoing debate on the right to explanation'' \cite{goodman2017european}. This debate has significantly influenced the shift in modern AI systems towards explainability. While the ultimate consumers of model explanations are humans, well-established research domains that focus on human understanding are underrepresented. For instance, only a few papers related to ``Cognition'' are cited compared to those on other algorithmic topics. Millecamp et al.~\cite{millecamp2019explain} suggest enhancing XAI theory with insights from social sciences, including cognitive science and psychology. Given the scant references to psychology, it appears that only a handful of XAI user studies delve into evaluating XAI from a psychological standpoint. We highlight a nascent research domain of XAI frameworks based on human cognition and behavior theories \cite{liao2021human}. This theoretical guidance can also offer conceptual tools for better evaluating XAI from user perspectives. More details about common references can be found in Appendix~\ref{appendix:foundation}.

\vspace{-9pt}
\subsection{Impact of User Studies}
\label{sec:impact}
\rev{\Cref{fig:combined} presents applications that make use (and thus are the consumers) of the findings from core papers. We noticed that studies on user understanding and trust span a wide range of applications. For example, trust is frequently addressed in the contexts of medical diagnosis and transportation, indicating its significance in high-risk scenarios.} Recommendation systems emerge as a primary focus in follow-up works. Papers on usability have a significant impact on fields like data visualization, software development, and education. In these areas, models frequently serve as tools to ease the burden on end users. Human-AI collaboration measures particularly promote the further development of robotics and or natural language processing. 
The prominence of recommendation systems in both foundational works and their impact implies that XAI is an integral component of contemporary recommendation systems. \rev{A comprehensive overview of the fundamental works and application domains can be found in Appendix~\Cref{appendix:combined}.}

\vspace{-5pt}
\section{Comprehensive User Study Analysis}
\label{sec:part2}
In this section, we present details of the covered XAI user studies. We first introduce some commonly used AI models and explanation techniques (\Cref{sec:models}), followed by a discussion of application domains and measures with respect to the four measured quantities. The experimental designs, as well as analysis tools are presented in \Cref{sec:analysis}.

\vspace{-5pt}
\subsection{Models and Explanations}
\label{sec:models}
As our selected core papers comprise a large spectrum of AI models, data modalities, and explanation approaches, we initially list the models and explanation techniques deployed along with the corresponding core paper references in \Cref{tab:models}.
It presents the utilization of explanation types in columns and model types in rows. The explanation methods used is organized according the the taxonomy by Molnar~\cite{molnar2019}. First, there are intrinsically interpretable models, also known as \textit{white-box models}. For instance, white-box models include decision trees and linear models.
Second, there are \textit{black-box models} that provide no parameter access or are too complex to be explained in a human-understandable way \cite{Rudin2019}. These include ensembling techniques such as Random Forests or neural models. 

As for explanation techniques, we identified five key types in the scope of the surveyed papers (rows of Table~\ref{tab:models}). Most frequently used are feature-based (attribution) explanations, for instance, SHAP (Shapley additive explanations \cite{lundberg2017unified}) and LIME (Local Interpretable Model-Agnostic Explanations \cite{ribeiro2016should}). 
There is a clear differentiation between local, instance-wise, explanations and global explanations that apply to the model in its entirety. For instance, the weights of a linear model have a global scope. 
This differentiation is common among these feature-based explanations, where most of the papers using local explanations. 
Other popular explanation types are example-based explanations, 
counterfactual explanations, which aim at providing actionable suggestions for attaining a user-preferred prediction by changing certain input features,
and concept-based explanations, which use meaningful high-level concepts such as objects or shapes to explain a prediction. 

\begin{table}[]
\centering
 \setlength{\belowcaptionskip}{-5mm}
\resizebox{\linewidth}{!}{
\begin{tabular}{cc|c|c|c}
\hline
\multicolumn{2}{c|}{}   & \textbf{White-box} & \textbf{Black-box} & \textbf{Other} \\ \hline
\multicolumn{1}{c|}{\multirow{2}{*}{\begin{tabular}[c]{@{}c@{}}\textbf{Feature-}\\ \textbf{based}\end{tabular}}} & \textbf{local}  & 
\begin{tabular}[c]{@{}c@{}} 
\cite{kaur2020interpreting, caruana2015intelligible, abdul2020cogam} \\ 
\cite{,cheng2019explaining, wang2021explanations, bell2022accuracyexplain} \\
\cite{tsai2021exploring, arora2022explain}
\end{tabular}     &  \begin{tabular}[c]{@{}c@{}}    
\cite{kaur2020interpreting,ramamurthy2020model,ribeiro2018anchors, plumb2020regularizing}\\
\cite{hadash2022improving,smith2020no,alufaisan2021does, rawal2020beyond} \\
\cite{bell2022accuracyexplain,chromik2021think,hase2020evaluating}  \\
\cite{bove2022contextualization,wang2022interpretableideation, bang2021explaining} \\
\cite{alqaraawi2020evaluating, chandrasekaran2018explanations, sixt2022do} \\
\cite{balayn2022can, nourani2021anchoring,arora2022explain}\\
\cite{hase2020evaluating, schuff2022human,bansal2021does} \\
\cite{lai2020chicago, lai2019human,colin2022what}\\
\cite{nguyen2021the, kim2022hive, shen2020useful}\\
\end{tabular}&
  \\ \cline{2-5} 
\multicolumn{1}{c|}{}        & \textbf{global} & 
\begin{tabular}[c]{@{}c@{}} 
\cite{wang2021explanations, poursabzi2021manipulating,dodge2019explaining} \\
\cite{kaur2020interpreting,arora2022explain}
\end{tabular}
&  \begin{tabular}[c]{@{}c@{}} 
\cite{arora2022explain}
\end{tabular}   &   \\ \hline
\multicolumn{2}{c|}{\textbf{Example-based}} &   
\begin{tabular}[c]{@{}c@{}} 
\cite{wang2021explanations,kaur2020interpreting, kim2022hive}\\
\cite{dodge2019explaining,tsai2021exploring,taesiri2022visual}
\end{tabular}
& 
\begin{tabular}[c]{@{}c@{}} 
\cite{borowski2021exemplary, cai2019effects, sixt2022do} \\
\cite{buccinca2020proxy,hase2020evaluating,lai2019human}
\end{tabular}
& \begin{tabular}[c]{@{}c@{}} \cite{ross2021evaluating} \\ (generative \\ models) \end{tabular} \\\hline
\multicolumn{2}{c|}{\textbf{Counterfactual}} &   
\begin{tabular}[c]{@{}c@{}}
\cite{wang2021explanations,kuhl2022keep} \\
\cite{kuhl2022let,kaur2020interpreting}

\end{tabular}
& \begin{tabular}[c]{@{}c@{}}
\cite{schoeffer2022there,zhang2018towards} \\
\cite{sixt2022do,wang2022interpretableideation}
\end{tabular}
&   \\ \hline
\multicolumn{2}{c|}{\textbf{Concept-based}} &  
&
\begin{tabular}[c]{@{}c@{}}
\cite{yeh2019completeness, ghorbani2019towards, balayn2022can} \\
\cite{zhang2022debiased, leemann2022coherence, laina2020quantifying}\\
\cite{sixt2022do,zhang2018towards}\\
\end{tabular}
&   \\ \hline
\multicolumn{2}{c|}{\textbf{Other}} 
&\begin{tabular}[c]{@{}c@{}}
\cite{schaffer2019can,das2020leveraging} \\
\cite{guo2022building,paleja2021utility}
\end{tabular}&
\begin{tabular}[c]{@{}c@{}}
\cite{panigutti2020doctor,panigutti2022understanding,suresh2022intuitively, zhang2020effect} \\ 
\cite{colley2021effects,buccinca2020proxy,antoran2021getting} \\
 \cite{colley2021effects,nourani2021anchoring, chandrasekaran2018explanations} \\
 \cite{ramamurthy2020model,ribeiro2018anchors,rebanal2021xalgo} \\
\cite{ehsan2019automated,le2018improving, dominguez2019effect} \\
 \cite{gao2019explainable,radensky2022exploring, rader2018explanations}* \\
 \cite{,wang2022humans, ooge2022explaining,kunkel2019let} * \\
 \cite{kouki2019personalized,tsai2019explaining}*
\end{tabular}
& \begin{tabular}[c]{@{}c@{}} 
\cite{anik2021data} \\
\cite{millecamp2019explain,li2019data,tsai2018beyond} $\dagger$ \\ 
\cite{rong2022user,choi2018fontmatcher,li2019data} $\dagger$ \\
\cite{schneider2021explain,arendt2020parallel,hohman2019gamut} $\dagger$ \\
\cite{le2018improving,smith2020digging, peng2022inherently} $\dagger$ \\
\cite{feng2019can,liao2021should} $\dagger$
\end{tabular} \\ \hline
\end{tabular}
}
\caption{Models and explanations in core papers. Papers are categorized according to types of explanations (\textbf{column}) and types of models (\textbf{row}). $*$ denotes papers using recommendation systems as models; $\dagger$ denotes papers proposing novel interpretable interfaces as studied models.}
\label{tab:models}
\end{table}

Besides these four main types of explanations, there are other explanations such as rules \cite{schaffer2019can,das2020leveraging} or game strategies \cite{guo2022building,paleja2021utility} when AI plays games. 
More details about concrete models and explanations can be found in Appendix~\ref{sec:modelexplanations}.

\vspace{-5pt}
\subsection{Measurements}
The effectiveness of explanations can be characterized from several angles. We specifically identified the categories of trust, understanding, usability, and human-AI collaboration performance. In this section, we give an overview of the contexts in which each of these variables is studied and the measures used to quantify them. 
\label{sec:measures}
\subsubsection{Trust}
User trust is studied in decision-making applications such as image classification \cite{buccinca2020proxy, cai2019effects}, (review) deception detection \cite{lai2019human} or loan approval \cite{schoeffer2022there}.
Besides decision making, \cite{dominguez2019effect, millecamp2019explain, liao2021should,kunkel2019let,ooge2022explaining,tsai2018beyond} study user trust in the domain of recommendation systems. 
Whether explainable ML models can increase user trust in the medical domain is studied in \cite{panigutti2022understanding, tsai2021exploring, suresh2022intuitively}. Moreover, Colley et al.~\cite{colley2021effects} measure user trust in an autonomous driving application with and without explanations. 



Trust measures used in much of the existing research can be divided into two groups: \textit{self-reported} and \textit{observed} trust \cite{papenmeier2019model}. Self-reported trust is commonly measured by asking users to fill out questionnaires whereas observed trust is quantified by humans' agreement with the model's decisions. In \Cref{tab:trust measures} in Appendix, trust measures in these two groups are listed. The agreement rate of users with the model decisions is commonly used \cite{suresh2022intuitively, wang2021explanations, schaffer2019can, lai2019human} as a measure of observed trust. 
Parallel to observed trust measurement, van der Waa et al.~\cite{van2021evaluating} ascribe the user's alignment behaviors to the \textit{persuasive power} of model explanations, i.e., the capacity to convince users to follow model decisions despite the correctness.
As an extension, trust calibration is defined based on this measure. For example, a high agreement rate to wrongly made decisions represents \textit{overtrust}, while a low agreement rate to correct decisions means \textit{undertrust} \cite{wang2021explanations}. 
In self-reported measurements, researchers either utilize well-developed questionnaires or self-designed ones, with the exception of \cite{ehsan2021expanding} which conducts a semi-structured interview to explore user opinions. Several works
\cite{tsai2021exploring, buccinca2020proxy, schoeffer2022there, schaffer2019can, dominguez2019effect, cai2019effects, millecamp2019explain, tsai2018beyond, kim2020answering} propose their own questionnaires. Among these, a subgroup \cite{dominguez2019effect, buccinca2020proxy, millecamp2019explain, tsai2018beyond, kim2020answering} simply asks users to rate a single statement such as ``I trust the system's recommendation/decision'', which is named as one-dimensional trust by \cite{ooge2022explaining}. When deploying previously proposed questionnaires \cite{anik2021data,colley2021effects,kaur2020interpreting,guo2022building,kunkel2019let,liao2021should,ooge2022explaining,paleja2021utility,cheng2019explaining,erickson2017machine}, Trust in Automation \cite{jian2000foundations} is the most commonly used one, in which the underlying constructs of trust between human and computerized systems are explored. 
\subsubsection{Understanding}
An important goal of explanation techniques is to foster users' understanding of complex ML systems.
An important separation has to be made between users' perceived understanding and their actual comprehension of the underlying model, as the two often do not agree \cite{chromik2021think, hase2020evaluating}. Cheng et al.~\cite{cheng2019explaining} explicitly differentiate between \textit{objective} understanding and self-reported understanding, which we term \textit{subjective} understanding in this work. While subjective understanding is usually measured through questionnaires, measuring objective understanding requires a proxy task where the users' understanding is put to a test. Additionally, user studies can be run to assess how well users can understand the explanation itself (and not the underlying model). This can be an important sanity check and is particularly used in the domain of conceptual explanations \cite{Kim2018interpretabilityTCAV, ghorbani2019towards}, where the intelligibility of concepts needs to be verified. We refer to the third category as \textit{understanding of explanations} but defer its detailed findings to Appendix~\ref{sec:understandingofexplanations}.

\textbf{Objective Understanding.} Works in the subdomain of objective understanding deploy proxy tasks to verify users' understanding of a model's inner workings. 
The most commonly considered domain in works on understanding is finance \cite{poursabzi2021manipulating, abdul2020cogam, ramamurthy2020model,hase2020evaluating, bove2022contextualization, chromik2021think, bell2022accuracyexplain} followed by image classification \cite{buccinca2020proxy, kaur2020interpreting, borowski2021exemplary}. 
One of the most critical design choices when assessing objective understanding is the selection of a suitable proxy task. Doshi-Velez and Kim~\cite{doshi2017towards} argue that the task should \textit {``maintain the essence of the target application''} that is anticipated. One of the most prominent tasks is forward simulation \cite{lipton2018mythos, doshi2017towards}. This task demands subjects that are given an input to simulate, i.e., predict, the model's output. The extent to which participants can successfully provide the model's output is also referred to as \textit{simulatability} \cite{lipton2018mythos}. However, scholars have designed many more tasks to quantify understanding and applied them across a variety of data modalities (cf. Table \ref{tab:proxytasks} in Appendix for an exhaustive listing). 

We briefly describe other common tasks below.
A special variant of forward simulation is called \textit{relative simulation}. In this task, users predict which example out of a predefined choice will have the highest prediction score (or class probability). \textit{A manipulation or counterfactual simulation task} \cite{doshi2017towards} asks users to manipulate the input features in such a way that a certain model outcome (counterfactual) is reached. Users' performance on this task can be used as a proxy for their understanding. Lipton~\cite{lipton2018mythos} pointed out that simulatability can only be a reasonable measure, if the model is simple enough to be captured by humans and that simpler tasks are required otherwise. 
An example could be a \textit{feature importance} query, where users have to tell which features are actually used by the model. A directed and more local version of this task is \textit{marginal effects queries}, where the subjects predict how changes in a given input feature will affect the prediction (e.g., \emph{``Does increasing feature $X$ lead to a higher prediction of $Y$ being class $1$?''}). Because explanations should allow the identification of weaknesses in models, the task of \textit{failure prediction} measures the accuracy of users' prediction when the model prediction is wrong.


\textbf{Subjective Understanding.}
Besides the objective understanding which is supported by performance indicators, understanding of a model may be subjective, i.e., it may depend on a user's own perception.
The most commonly used applications that measure subjective understanding are various recommendation system setups \cite{radensky2022exploring, hadash2022improving, dominguez2019effect, rader2018explanations}.

Most of the works assess the subjective understanding of a user with a post-task questionnaire. 
Guo et al.~\cite{guo2022building} adapted a popular questionnaire designed for recommendation systems by Knijnenburg et al.~\cite{knijnenburg2012explaining}, while Bell
et al.~\cite{bell2022accuracyexplain} accommodated the questionnaire which originally intended to measure the intelligibility of differenet explanations by Lim and Dey~\cite{lim2009assessing}. 
On the other hand, agreement to simple subjective statements such as \textit{``I understand this decision algorithm''} \cite{cheng2019explaining}, \textit{``I understand how the AI...''} \cite{buccinca2020proxy, cai2019effects} or \textit{``The explanation(s) help me to understand...''} \cite{radensky2022exploring} can be collected to assess subjective understanding. 

\subsubsection{Usability}
Usability is a key concern of every HCI system and thus applies to almost all domains.
This is reflected in the surveyed papers, where usability is studied in a wide range of setups and contexts. We also include application-specific performance measures in this category. 
Based on the measurements in the user studies, we refined usability into measures of helpfulness, workload (cognitive load), satisfaction, ease of use and \revise{detecting undesired behaviors of the system}, as shown in \Cref{tab:categories}. 
To assess workload (cognitive load), NASA-TLX scale \cite{hart1988development} is used in \cite{colley2021effects, kaur2020interpreting, dominguez2019effect,tsai2021exploring,arendt2020parallel}, while Abdul et al.~\cite{abdul2020cogam} measure cognitive load by capturing the log-reading time of memorizing the explanation. 
Most of the works use self-designed questionnaires or statements to measure satisfaction \cite{millecamp2019explain, tsai2018beyond, tsai2019explaining, tsai2021exploring, dominguez2019effect, smith2020digging, smith2020no, kouki2019personalized}, however, the Explanation Satisfaction Scale \cite{hoffman2018metrics} can be deployed as an established alternative \cite{bove2022contextualization, panigutti2022understanding}.
Helpfulness can be assessed by simply asking for subjective ratings of the explanations for accomplishing a specific task \cite{gao2019explainable,buccinca2020proxy,nourani2021anchoring,wang2022interpretableideation,zhang2022debiased,zhang2022towards}. 
Colley et al.~\cite{colley2021effects} use an adapted version of the System Usability Scale proposed in \cite{holzinger2020measuring}. 

\rev{Using model explanations to audit models is one purpose of explainability~\cite{arrieta2020explainable}. Some of the surveyed works study how model explanations can assist users in detecting undesired behaviors of models. These issues mainly include (perceived) unfairness in the model decision-making~\cite{rader2018explanations,dodge2019explaining, binns2018s,schoeffer2021appropriate}, biases in models~\cite{rawal2020beyond} or features~\cite{sixt2022do}, and wrong decisions (failures)~\cite{kim2020answering} in the studied papers. A detailed summary of types of undesired behaviors is listed in~\Cref{tab:undesired behavior detection}.}
In the undesired behavior detection, the effectiveness of explanations is evaluated by objective performance measures, such as the number of bugs identified \cite{balayn2022can}, the share of participants that identify a certain bias \cite[First Experiment]{sixt2022do}
or by the deviations between model predictions and human predictions for unusual samples \cite{poursabzi2021manipulating}.  
\revise{The perception of users regarding fair treatment by a system has primarily been researched in high-stakes applications such as granting loans \cite{schoeffer2022there} or granting bail for criminal offenders \cite{dodge2019explaining, harrison2020empirical, grgic2018human}. For example, \cite{dodge2019explaining, harrison2020empirical, grgic2018human} investigate the fairness of COMPAS, a commercial criminal risk estimation tool that was used in the US to help make judicial bail decisions.}
It is also considered in everyday use-cases such as news \cite{rader2018explanations} and music \cite{htun2021perception} recommendations, or possible career suggestions \cite{wang2022humans}, where a bias in the underlying system can be to the detriment of the user. 
As the assessment of fairness is a very subjective matter, questions regarding perceived fairness are prevalent, 
e.g., ``how the software made the prediction was fair'' \cite{dodge2019explaining}, which can be answered on 5- or 7-point Likert scales \cite{dodge2019explaining, harrison2020empirical, schoeffer2022there, rader2018explanations, anik2021data, grgic2018human}. \rev{Among these works, an effective explanation is the one that can either increase or decrease the fairness perceptions,  since the aim of explanations is to show fairness or unfairness.} An exhaustive overview of measures for usability is given in \Cref{tab:usability measures} of the Appendix.
\vspace{-5pt}
\subsubsection{\revise{Human-AI Collaboration Performance}}
The goal of human-AI teaming is to improve the performance in AI-supported decision-making above the bar set by humans or an AI alone \cite{bansal2021does}. Improving human performance with the help of AI has been considered in games \cite{paleja2021utility, das2020leveraging}, question answering tasks \cite{feng2019can, bansal2021does}, deception detection \cite{lai2019human, lai2020chicago} and topic modeling \cite{smith2020digging, smith2020no}. 



The most common assessment is to rate AI-aided human performance by the percentage of correctly predicted instances in the decision-making process \cite{lai2019human, lai2020chicago,bansal2021does}. Paleja et al.~\cite{paleja2021utility}, however, define the performance as the time to complete the task. In \cite{das2020leveraging}, performance is measured in a game-based application, chess, using a winning percentage (which is commonly used in sports) as well as a percentile rank of player moves. 

\vspace{-6pt}
\subsection{Experimental Design and Analysis}
\label{sec:analysis}
There are three common experimental settings when conducting user evaluation: between-subjects (or between-groups) designs, within-subjects designs, and mixed designs that combine elements of both. An overview of the designs found in the core papers and their participant numbers is presented in \Cref{tab:experimentaldesign} and \Cref{fig:user statistics}, respectively.

\begin{table}[b]
    \centering
    \setlength{\belowcaptionskip}{-4mm}
    \resizebox{\linewidth}{!}{
    \begin{tabular}{cccc}\hline
      & \multicolumn{3}{c}{\textbf{Experimental Design}} \\ \hline
        & Between-Subjects & 
        Within-Subjects & 
        Mixed 
        \\ \hline
        \textbf{Papers} & 
        \begin{tabular}[c]{@{}c@{}} \cite{liao2021should,guo2022building,ooge2022explaining, wang2021explanations,schoeffer2022there, colin2022what, zhang2020effect} \\\cite{cai2019effects,kaur2020interpreting,cheng2019explaining,kunkel2019let,lai2019human, rawal2020beyond}
        \\ \cite{ehsan2019automated, schaffer2019can, hase2020evaluating, zhang2022towards, ross2021evaluating, nguyen2021the} \\
        \cite{poursabzi2021manipulating, bove2022contextualization, ramamurthy2020model, arora2022explain, antoran2021getting, choi2018fontmatcher} \\
        \cite{nourani2021anchoring, bell2022accuracyexplain, rader2018explanations, kuhl2022keep, rebanal2021xalgo, kim2022hive} \\
        \cite{lai2020chicago, alqaraawi2020evaluating, smith2020digging, smith2020no, alufaisan2021does, taesiri2022visual}
        \\ \cite{sixt2022do, chandrasekaran2018explanations, wang2021explanations, grgic2018human, rader2018explanations}\\
        \cite{harrison2020empirical, wang2022humans, htun2021perception, binns2018s, kuhl2022let}
        \end{tabular} & 
        \begin{tabular}[c]{@{}c@{}} \cite{panigutti2022understanding,colley2021effects,ehsan2021expanding,tsai2018beyond,suresh2022intuitively} \\ \cite{millecamp2019explain,tsai2019explaining,paleja2021utility,kim2020answering,kaur2020interpreting} 
        \\ \cite{rong2022user, buccinca2020proxy, borowski2021exemplary, chromik2021think, feng2019can}
        \\ \cite{sixt2022do, binns2018s, balayn2022can, hohman2019gamut, gajos2022people} \\
        \cite{kouki2019personalized, tsai2021exploring, zhang2022debiased, leemann2022coherence, ghorbani2019towards}\\
        \cite{gao2019explainable, schuff2022human, shen2020useful, peng2022inherently, plumb2020regularizing}
        \end{tabular} &\begin{tabular}[c]{@{}c@{}} \cite{anik2021data,arendt2020parallel, buccinca2020proxy,borowski2021exemplary,dominguez2019effect} \\
        \cite{dodge2019explaining,ehsan2019automated,hadash2022improving, laina2020quantifying,paleja2021utility}\\
        \cite{radensky2022exploring, ribeiro2016should, schneider2021explain, wang2022interpretableideation, wang2021explanations}\\
        \cite{yeh2019completeness} \end{tabular} \\\hline
      \end{tabular}
      }
      \caption{Experimental designs in core papers.}
      \label{tab:experimentaldesign}
\end{table}

\vspace{-6pt}
\subsubsection{Between-subjects}
With slightly above \revise{$55\,\%$} of the user studies conducted in a between-subjects manner, i.e., one subject is only exposed to one condition, this design choice is most common in the XAI literature. The number of participants in the between-subjects manner usually starts at around 30 participants, while it may go up to 1070 in total for 3 conditions as in \cite{cai2019effects} and to 1250 for 5 conditions in \cite{poursabzi2021manipulating}. \revise{However, the number of participants can be limited when the studied application is designed for specific groups of lay persons, which cannot be easily recruited from the Internet platforms such as Amazon Mechanical Turk. For instance, Ooge et al.~\cite{ooge2022explaining} use 12 school students per condition.} Some authors place particular emphasis on participants being similar to the average demographic \cite{harrison2020empirical, grgic2018human}.

The conditions usually include the different explanation techniques in combination with other parameters such as the model, data set, data modality, or a number of features used as independent variables. Note that a full grid design with many independent variables may quickly result in a very high number of conditions, 
which in turn requires many participants. The outcome variable of interest is commonly measured on a numerical or ordinal scale right away, however, in the fairness domain, qualitative analyses are sometimes obtained through conducted interviews or written responses \cite{anik2021data, grgic2018human, schoeffer2022there}.

The statistical analysis directly follows from this design. If one is interested in identifying significant differences between the groups, common statistical hypotheses tests are used. For overall comparison, one or two-way ANOVA tests are the most commonly used statistical tool. Interesting post-hoc comparisons between two groups can be made with a standard T-test, if the data is normally distributed with equal variance, or by using non-parametric tests such as the Wilcoxon rank-sum test (also known as Mann-Whitney U-test) for comparison of two populations (e.g, \cite{sixt2022do}) or the Tukey HSD test (e.g., \cite{ramamurthy2020model}) for multiple populations. When running multiple post-hoc tests, some works make use of the Bonferroni correction (e.g, \cite{sixt2022do}).

\begin{figure}[t]
    \centering
     \setlength{\belowcaptionskip}{-5mm}
     \setlength{\abovecaptionskip}{-4mm}
    \includegraphics[width=.7\linewidth]{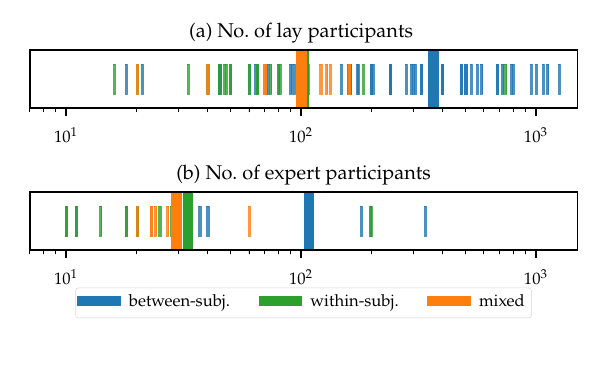}
    \caption{Distribution of participant numbers in the surveyed user studies by design and participant type (each bar represents one study). Per-design means are indicated in bold.}
    \label{fig:user statistics}
\end{figure}

\vspace{-6pt}
\subsubsection{Within-subjects}

Around \revise{$30\,\%$} of the papers use the within-subjects design, where each participant sequentially passes through all conditions and provides feedback.
Fewer participants are recruited in within-subjects experiments compared to the between-subjects ones. Hence, they are particularly popular when participants with restrictive characteristics, such as domain-specific professional expertise, are required. For example, Suresh et al.~\cite{suresh2022intuitively} and Rong et al.~\cite{rong2022user} recruit fourteen medical professionals and five radiologists in their user studies, respectively. The small number of medical experts contributing to the user study is a limitation \cite{rong2022user}, however, it is often the case in expert user research. Gegenfurtner et al.~\cite{gegenfurtner2011expertise} evaluate $73$ sources and point out that the majority of these studies include only five, maybe ten experts. Besides the medical domain, other works \cite{ehsan2021expanding,colley2021effects,tsai2018beyond,kaur2020interpreting} also invite subjects with particular professions such as engineers in a technology company. When no specific knowledge is required, however, participant numbers reach up to $740$ also for within-subjects designs \cite{gajos2022people}. 

For within-groups designs, the Wilcoxon signed-rank test (e.g. used by \cite{borowski2021exemplary, chromik2021think}) is the most common method to compare paired samples for significant differences. Repeated-measures ANOVA is a common analysis tool, when multiple comparisons are required (see, e.g., \cite{chromik2021think}).

\vspace{-6pt}
\subsubsection{Mixed}
The smallest group of studies, about \revise{$15\%$}, use a mixture of between- and within-subjects settings. In these works, subjects are first assigned randomly to one group, where they are exposed to multiple conditions. Anik and Bunt~\cite{anik2021data} use knowledge background in machine learning as a between-subjects factor to divide the participants into three groups (expert, intermediate and beginner), while inside each group participants interact with explanations in the context of four different scenarios (e.g., facial expression recognition or automated speech recognition). Dominguez et al.~\cite{dominguez2019effect} make the presence of explanations a between-subjects condition and different types of explanations a within-subjects factor in the group with model explanations. A particular challenge for such a study design is that statistical tools from both the independent-samples and dependent-samples categories need to be combined.

\vspace{-6pt}
\section{Findings of User studies}
\label{sec:previous findings}
In this section, we summarize the primary findings from the core papers.
\Cref{tab:findings} lists findings with respect to four measured quantities. To build an overview of the findings, we divide papers according to their evaluation dimensions, i.e., the independent variables in the user studies. When using the presence of explanations as the evaluation aspect, the findings are summarized in \Cref{tab:findings}. The listed impacts using explanations are to be seen in comparison with a control group without explanations. Effects are divided into two groups: (1) Positive effects, for example, increasing user trust or understanding; (2) Non-positive effects: the effect can be negative, or not significantly positive (neural), or a mixture of different effects (e.g., feature-based explanations have positive effects but counterfactual explanations do not). Beyond the explanations themselves, other possible evaluation dimensions such as that might have an impact on the perception of XAI, for instance, AI technology literacy, model performance, or the dimensionality of the data. Instead of using the mere presence of explanations, many works compare different explanation techniques with each other (see Appendix~\ref{sec:appfindings} for more details).

As various research questions and findings are addressed in \numcorepapers core papers, many papers compare explanation types in order to choose a preferable one, it is not possible to cover all results in one table. Based on them, we outline some interesting trends in the effectiveness of explanations on user experience: (1) Explanations are effective in improving users' subjective understanding; (2) The effectiveness of explanations in increasing user trust and usability of models is not clear; (3) Explanations are not good at convincing users that models are fair; (4) Interactivity of the model has positive impact on user trust, understanding and model usability. The first three statements can validated through the number of papers obtaining positive or non-positive effects in each category, while the last finding is extracted from \Cref{tab:findings2} in the Appendix, which details findings with on other independent variables. We encourage the reader to consider the short summary of \textit{primary} findings in the tables and check for further details according to their specific interests. In the following section, we highlight some findings for each category of measurement. 

\begin{table*}[h]
 \setlength{\belowcaptionskip}{-5mm}
\resizebox{\linewidth}{!}{
\begin{tabular}{cc|cc}
\specialrule{.1em}{.05em}{.05em}

\multicolumn{2}{c|}{\multirow{2}{*}{}} & \multicolumn{2}{c}{\textbf{Evaluation Dimension: Explanations}}                        \\ 
\multicolumn{2}{c|}{\multirow{2}{*}{}}   & \multicolumn{2}{c}{Effect of explanations compared to \textbf{no explanations}}    
\\\cline{3-4}
\multicolumn{2}{c|}{}    & \multicolumn{1}{c|}{Positive} & \multicolumn{1}{c}{{Non-positive / Mixed}} \\ \specialrule{.1em}{.05em}{.05em}
\multicolumn{2}{c|}{\textbf{Trust}}  & \multicolumn{1}{c|}{ \begin{tabular}[c]{@{}c@{}} \cite{buccinca2020proxy}: example-based, rule-based explanations 
\\ 
\cite{dominguez2019effect}: example-based explanations for recommendations \\ \cite{schoeffer2022there}: feature importance 
\\ \cite{paleja2021utility}: decision-tree explanation for policy \\
\cite{ehsan2019automated}: explanation corpus given by researchers \\
\cite{lai2019human}: feature-based (saliency map), example-based explanations \\
\cite{ooge2022explaining}: explanations for recommendations \\
\cite{tsai2021exploring}: rationale-based, example-based and feature-based (best) \\ explanations for online symptom checkers \\
\revise{\cite{zhang2020effect}: confidence scores}
\end{tabular}}    
&  \begin{tabular}[c]{@{}c@{}} \cite{colley2021effects}: positive in simulation \\ but no improvement in real-word \\
\cite{panigutti2022understanding}: 
explanations for medical suggestions (Doctor XAI \cite{panigutti2020doctor}) \\ pos.for observed trust but insignificant  for reported trust \\
\cite{wang2021explanations}: feature-based explanations increase appropriate trust \\ slightly but counterfactual explanations inconclusively \\
\cite{cheng2019explaining,kaur2020interpreting,kim2020answering}: feature-based explanation, insignificant \\
\cite{schaffer2019can}: rule-based explanation, insignificant \\ 
\revise{\cite{zhang2020effect}: Shapley values, insignificant} \\
\underline{\cite{smith2020no}: feature-based explanation, negative} \\ 
\end{tabular} 
\\ \specialrule{.1em}{.05em}{.05em}

\multicolumn{1}{c|}{\multirow{2}{*}{\textbf{Understanding}}}                                      & Obj. &  \multicolumn{1}{c|}{\begin{tabular}[c]{@{}c@{}}
\cite{poursabzi2021manipulating, cheng2019explaining} white-box model \\
\cite{hase2020evaluating} feature importance, LIME (tabular)\\
\cite{zhang2022towards} counterfactuals+cues (audio)\\
\cite{arora2022explain} manipulatability improved by white-box log. reg.\\
\cite{alqaraawi2020evaluating} saliency maps (image)\\
\cite{colin2022what} saliency maps for bias detection and strategy identification\\
\cite{wang2021explanations} counterfactuals+feature importance\\
\end{tabular}
}     
&  
\begin{tabular}[c]{@{}c@{}}
\underline{\cite{bell2022accuracyexplain}: SHAP, negative for black-box model (education domain)} \\
\cite{bell2022accuracyexplain}: Insignificant difference btw. black-box and white-box models\\
\cite{hase2020evaluating}: Prototypes, Anchors, LIME on textual data insignificant \\
\cite{zhang2022towards}: Counterfactuals and Concepts insignificant (audio data) \\
\cite{arora2022explain}: Simulatability results insignificant for LIME,\\ IG, surrogate model on BERT and Logistic Regression Model,\\ Manipulatability insignificant for BERT\\
\cite{chandrasekaran2018explanations}: Insignificant results for GRAD-CAM, \\Saliency Map, uncertainty scores in VQA\\
\revise{\cite{colin2022what} saliency maps for failure prediction (image)}\\
\revise{\underline{\cite{shen2020useful}: saliency maps, negative for a mix of three interpretation techniques in simulation task}}
\end{tabular}
\\ \cline{2-4} 
\multicolumn{1}{c|}{}
& Sub. & \multicolumn{1}{c|}{\begin{tabular}[c]{@{}c@{}} 
\cite{buccinca2020proxy}: example-based, rule-based explanations\\
\cite{ehsan2019automated}: explanation corpus given by researchers \\
\cite{wang2021explanations}: feature-, example- and counterfactual-based \\
\cite{rader2018explanations}: explanations provided by \cite{cotter2017explaining} for Facebook News Feed \\
\cite{dominguez2019effect}: example- and feature-based explanations \\
\cite{cai2019effects}: example-based explanations \\
\cite{hadash2022improving}: feature importance, SHAP and LIME \\
\cite{chromik2021think}: feature importance, SHAP \\
\end{tabular}}     
&  \multicolumn{1}{c}{\begin{tabular}[c]{@{}c@{}} 
\cite{cheng2019explaining}: white-box model, insignificant \\
\cite{bell2022accuracyexplain}: 
white-box < black-box, both insignificant \\
\revise{\underline{\cite{springer2019progressive}: feature importance explanation (transparent system) can be distracting}}
\end{tabular}}  
\\ \specialrule{.1em}{.05em}{.05em}
\multicolumn{2}{c|}{\textbf{Usability}}        & 
\multicolumn{1}{c|}{\begin{tabular}[c]{@{}c@{}} 
    \cite{hohman2019gamut}: counterfactuals, pos. for usability \\
    \cite{dominguez2019effect, bove2022contextualization}: example-based explanations, pos. for satisfaction \\
    \cite{zhang2022debiased}: CAM-related explanations, pos. for helpfulness \\
    \cite{tsai2021exploring}: rational-, feature-, example-based explanations,\\ pos. for satisfaction \\
    \cite{tsai2019explaining}: content-based explanations, pos. for satisfaction \\
    \cite{schneider2021explain}: explanations regarding driving information, \\pos. for ease of use\\
    \cite{buccinca2020proxy}: example-based and rule-based explanations, \\ pos. for helpfulness \\
    \cite{balayn2022can}: local, global, visual (saliency map) explanations, \\
    pos. for bug identification \\
    \cite{wang2022interpretableideation}: attribution methods and conceptual explanations,\\ pos. for usefulness\\
    \cite{choi2018fontmatcher}: feature-based, pos. for reliability \\
    \cite{kim2020answering}: (proposed) template-based expl. \\ pos. for debugging and usefulness \\
\cite{schoeffer2022there}: feature importance, counterfactual explanations \\ pos. for perceived fairness 
\end{tabular}}     
&  
\multicolumn{1}{c}{\begin{tabular}[c]{@{}c@{}} 
    \cite{kuhl2022let}: counterfactuals, significant for helpfulness/usability \\ but insignificant for usefulness \\
    \cite{panigutti2022understanding}: ontology-based explanation, insignificant for satisfaction \\
    \cite{wang2022interpretableideation}: attribution methods and conceptual explanations, \\ insignificant for ease of use \\
    \cite{kim2020answering}: visual explanations increases usefulness, \\ but
    improvement is insignificant \\
    \cite{colley2021effects}: pos. for cognitive load/usability (simulation), \\ but insignificant in real-world \\
    \underline{\cite{smith2020no}: feature-based explanations, negative for satisfaction} \\
\cite{rader2018explanations}: \underline{informing users about the algorithmic decisions, negative} \\
ranking scores of recommendations, insignificant for perceived fairness\\
\cite{schoeffer2022there}: highlight features only, insignificant for perceived fairness \\
\cite{binns2018s}: insignificant in between-subjects\\ but significant in within-subjects for perceived fairness
\end{tabular}}  
\\ \specialrule{.1em}{.05em}{.05em}
\multicolumn{2}{c|}{\begin{tabular}[c]{@{}c@{}}\revise{\textbf{Human-AI}}\\ \revise{\textbf{Collaboration Performance}}\end{tabular}} &   
\multicolumn{1}{c|}{\begin{tabular}[c]{@{}c@{}} 
    \cite{das2020leveraging}: textual explanations with domain knowledge  (in chess) \\
    \cite{lai2020chicago,lai2019human}: feature-based explanations \\
    \cite{feng2019can}: exampled-based for experts, feature-based for novices\\
    \cite{gajos2022people}: contrastive explanations \\
    \cite{buccinca2020proxy}: example-based and rule-based explanations\\
    \revise{\cite{nguyen2021the, taesiri2022visual}: example-based explanations, attributions (AI correctness prediction)} \\
    \revise{\cite{nguyen2022visual}: important parts in images as explanations}
    \end{tabular}}   
&  \multicolumn{1}{c}{\begin{tabular}[c]{@{}c@{}} 
 \cite{lai2019human}: exampled-based, insignificant \\
 \revise{\cite{bansal2021does, zhang2020effect}}: feature-based explanations, insignificant \\
\end{tabular}}    
\\ \specialrule{.1em}{.05em}{.05em}
 \end{tabular}}
\caption{User study findings when using model \textbf{explanations} as evaluation dimensions. Effects of explanations compared to the baseline (control group) of ``no explanations'' on measured quantities. Effects are divided into ``Positive'' where explanation information is given, and ``Non-positive / Mixed'' where negative impact is marked with \underline{underlines}.}
\label{tab:findings}
\end{table*}

\vspace{-5pt}
\paragraph*{Trust} Among the papers comparing the effect of using explanations to using no explanations, or placebo (randomly generated) explanations \cite{lai2019human,ooge2022explaining}, about half of the papers validate that explanations have a positive impact on user trust \cite{panigutti2022understanding,ooge2022explaining, buccinca2020proxy, dominguez2019effect,schoeffer2022there, lai2019human, paleja2021utility,ehsan2019automated}, while the other half cannot verify this hypothesis \cite{colley2021effects, wang2021explanations,schaffer2019can,kaur2020interpreting,cheng2019explaining,kim2020answering}. 
For instance, Colley et al.~\cite{colley2021effects} investigated the explanations in an autonomous driving task and discover that the trust is improved in simulation but not with the real-world footage. Another example of the mixed effect of using explanations is found in \cite{wang2021explanations}, where (minimal) evidence is found that feature-based explanations help increase appropriate trust, but counterfactual explanations do not. 

Apart from using explanations as independent variables, the user personalities or expertise may also affect their perceptions 
\cite{anik2021data, cai2019effects, cheng2019explaining, kunkel2019let, millecamp2019explain,smith2020digging}. 
Millecamp et al.~\cite{millecamp2019explain} captured personal characteristics in the aspects such as the Locus of Control defined by Fourier (``the extent to which people believe they have power over events in their lives''), Need for Cognition (``a measure of the tendency for an individual to engage in effortful cognitive activities'') or Tech-Savviness (``the confidence in trying out new technology''). However, no significant interaction effect could be found between the personal characteristics and the trust. Liao and Sundar~\cite{liao2021should} studied a recommendation system asking users’ personal data with different explanations. They hypothesized that explanations in a ``help-seeker'' style and using the pronoun ``I'' would gain more trust of users than the explanations formalized in a ``help-provider'' style. Nevertheless, However, the opposite result is found and using self-referential expression resulted in lower affective trust. Model performance together with model explanation was studied in \cite{cai2019effects} for an image recognition task. The authors found out when images were recognized (high model performance), users feel the system more capable (``capability'' is defined as a belief of trust).

\vspace{-4pt}
\paragraph*{Understanding}
The fundamental question in this subdomain is to find out which explanation technique is most beneficial for increasing the user's understanding of a machine learning model. As pointed out earlier, understanding can be measured both in a subjective and objective manner. 
 
We first discuss results on objective understanding. The goal of increasing objective understanding was explicitly posed by Alqaraawi et al.~\cite{alqaraawi2020evaluating} who reported that saliency maps have a positive effect on understanding. Wang
and Yin~\cite{wang2021explanations} show that counterfactual explanations and feature importance increase users objective understanding. On the contrary, Sixt et al.~\cite{sixt2022do} find none of their examined explanation techniques (counterfactuals, conceptual explanations) superior to a baseline technique consisting of example images for each class and the work by Hase and Bansal~\cite{hase2020evaluating} reveals that many explanations (including anchors, prototypes) have no effect in increasing objective understanding, which LIME on tabular data being the only exception. 
Apart from the explanation, several other factors have been identified to have an effect on objective understanding. Hase and Bansal~\cite{hase2020evaluating} suggest that the \textit{data modality} may have a non-negligible impact on how different explanation techniques increase understanding. 
Some results highlight that the \textit{choice of proxy task} is influential. Arora et al.~\cite{arora2022explain} show that their manipulatablity task revealed differences remained hidden when forward simulation is used. In spite of these findings, Buçinca et al.~\cite{buccinca2020proxy} underline that preferred explanations may be different in a real-world application from a simulated one. Regarding the \textit{type of model}, there is disagreement on whether white or black-box models can lead to increased objective understanding. While black-box models without explanations resulted in higher simulation performance than white-box models with SHAP values in \cite{bell2022accuracyexplain}, Cheng et al.~\cite{cheng2019explaining} observe that white-box models increase simulatability and also conclude that \textit{interactivity} is an important factor when it comes to objective understanding. 

In comparison with the objective understanding, the research question in the subdomain subjective understanding is to find out how explanations impact user's \textit{perceived} understanding \cite{radensky2022exploring, guo2022building, kuhl2022keep, ross2021evaluating, wang2021explanations, cheng2019explaining, nourani2021anchoring, cai2019effects, hadash2022improving}.
There exist a trend of using model explanations to improve subjective understanding \cite{buccinca2020proxy,cai2019effects,dominguez2019effect,ehsan2019automated,hadash2022improving,rader2018explanations,wang2019designing}. However, Chromik et al.~\cite{chromik2021think} challenge the improvement in perceived understanding with the cognitive bias named \textit{illusion of explanatory depth} (IOED) \cite{rozenblit2002misunderstood}, which means that laypeople often have overconfidence bias in their understanding of complex systems. Their results confirm the IOED issue in XAI, i.e., questioning users' understanding by asking them to apply their understanding in practice consistently reduces their subjective understanding. Explanations can have different impacts on subjective and objective understandings \cite{cheng2019explaining}, where white-box explanations increase objective understanding but do not have significant impact on subjective understanding. Similar disagreements have been observed in multiple other works \cite{hase2020evaluating, wang2019designing}. Radensky et al.~\cite{radensky2022exploring} examine the joint effects of local and global explanations in a recommendation system and their results provide evidence that both are better than either alone.

\vspace{-6pt}
\paragraph*{Usability} Similar to trust, it is not clear whether explanations are effective in improving users' perceptions of helpfulness, satisfaction or other dimensions of usability. For instance, in \cite{dominguez2019effect, bove2022contextualization, smith2020digging}, the explanations have a positive effect on satisfaction, while no significant effects on satisfaction are observed in \cite{kouki2019personalized, tsai2018beyond, millecamp2019explain,smith2020no}. Parallel to trust, Smith-Renner et al.~\cite{smith2020no} provide evidence for the hypothesis that it is harmful to user trust and satisfaction to show explanations by highlighting the important words in a text classification task. A strong correlation between self-reported trust and satisfaction can also be observed in \cite{colley2021effects}, where explanations have a positive impact in a simulated driving environment, but no significant effects when using real-world data. Beyond explanations, Nourani et al.~\cite{nourani2021anchoring} study the order of observing system weakness and strengths, which reveals that encountering weakness first results in a lower rate of usage of system explanations than encountering strength first. 
Schoeffer et al.~\cite{schoeffer2022there} find out that showing feature importance scores or counterfactual explanations (or a combination of both) for explaining decisions helps increase the perceived fairness, whereas highlighting important features without scores does not. However, several studies don't show a significant difference between scenarios with and without explanations~\cite{rader2018explanations, schoeffer2022there, binns2018s}.
Effects of explanations may be dependent on input samples, as shown in \cite{zhang2022debiased}. The authors show that both Debiased-CAM and Biased-CAM improve the helpfulness for a weakly blurred image, however, there is no significant improvement for unblurred or strongly blurred images. \rev{When used to assist users in detecting undesired behaviors,  model explanations are likely to identify various types of problems that exist within models or data, as demonstrated by \cite{rawal2020beyond,sixt2022do, balayn2022can}. However, successful detection is not guaranteed. For example, Poursabzi-Sangdeh et al. \cite{poursabzi2021manipulating} show that users with model explanations are less able to identify incorrect predictions. A limitation of current detection methods is that users may have varying assessments, such as perceived unfairness and irrelevance~\cite{grgic2018human,poursabzi2021manipulating,balayn2022can}, regarding the features used in models for decision-making. Due to this limitation, the effectiveness of methods assessed through self-reported data may face challenges in generalizability as discussed in~\cite{grgic2018human}. Yet, these methods generally offer a \textit{one-size-fits-all} solution, failing to account for variations in individual assessments.}

\begin{table}[t]
\resizebox{\linewidth}{!}{
\begin{tabular}{c|c|c}
\specialrule{.1em}{.05em}{.05em}
 \textbf{Type} & \textbf{Paper}
& \textbf{Detection Result}
\\
\specialrule{.1em}{.05em}{.05em}
\begin{tabular}[c]{@{}c@{}} Wrong decisions\\ (failures) \end{tabular} & \cite{balayn2022can,poursabzi2021manipulating,kim2020answering} & \begin{tabular}[c]{@{}c@{}}High detection rate in~\cite{kim2020answering}; \\ Moderate detection rate (50\%) in~\cite{balayn2022can};\\   Lower detection rate in ~\cite{poursabzi2021manipulating} \end{tabular}\\\hline

\begin{tabular}[c]{@{}c@{}} Biases in features \\ used by models\end{tabular} & \cite{sixt2022do, balayn2022can} &   \begin{tabular}[c]{@{}c@{}}Moderate detection rate (50\%) in~\cite{balayn2022can}; \\ Moderately high detection rate (>50\%)\end{tabular}\\\hline

\begin{tabular}[c]{@{}c@{}} Discrimination/Biases \\ in decisions \end{tabular} & \cite{rawal2020beyond} &  \begin{tabular}[c]{@{}c@{}} Humans perform well in \\ bias detection (accuracy=88.9\%) and \\ bias description (66.7\%) \end{tabular}\\\hline 

\begin{tabular}[c]{@{}c@{}} Unfairness (perceived) \\ in models \end{tabular} & \begin{tabular}[c]{@{}c@{}} \cite{schoeffer2022there,rader2018explanations, anik2021data, dodge2019explaining} \\ \cite{binns2018s,schoeffer2021appropriate,grgic2018human} \end{tabular} 
 & \begin{tabular}[c]{@{}c@{}} Succeed to judge~\cite{binns2018s,schoeffer2022there,dodge2019explaining}; \\ Not succeed to judge~\cite{anik2021data}; \\ Not always (no consensus) ~\cite{rader2018explanations,grgic2018human}\end{tabular} \\\hline
\specialrule{.1em}{.05em}{.05em}
\end{tabular}
}
\caption{\rev{Overview of results for undesired behavior detection using model explanations.}}
\vspace{-10pt}
\label{tab:undesired behavior detection}
\end{table}

\vspace{-5pt}
\paragraph*{\revise{Human-AI Collaboration Performance}}
A strain of works \cite{das2020leveraging, feng2019can, lai2019human, lai2020chicago, nguyen2021the, taesiri2022visual, nguyen2022visual} show that viewing explanations can improve human accuracy in making decisions, especially with feature-based explanations taking text data as input \cite{lai2019human,lai2020chicago,feng2019can}. When using example-based explanations in text classification, there is no improvement in human performance \cite{lai2019human}. Likewise, utilizing explanations has no significant impact on human performance in \cite{alufaisan2021does,bansal2021does}, but simply showing model predictions has a positive effect in \cite{alufaisan2021does}. Experts and novices perceive explanations differently, for example, Feng and Boyd-Graber~\cite{feng2019can} conclude that the performance gain of novices and experts comes from different explanation sources. Paleja et al.~\cite{paleja2021utility} reveal that explanations can improve novices' performance but decrease experts' performance. Additionally, less complex models with explanations can better convince humans in correct decisions \cite{lai2020chicago}.

\vspace{-7pt}
\section{A Guideline for XAI User Study Design}
\label{sec:part3}
\vspace{-2pt}
Learning from the best practices of the previous works, we summarize a handy guideline for XAI user study, which serves as a checklist for XAI practitioners. This guideline contains suggestions to avoid pitfalls that researchers could easily overlook. We introduce our guidelines in the order of before, during and after user studies, which reflects user study design, execution and data analysis, respectively. 

\vspace{-5pt}
\paragraph*{Before the User Study} When designing a user study, the first step is to decide what to measure. To define the measured quantities, one can consider two alternatives: using a general definition or an application-based quantity that is specific to the application at hand. The former one refers to a quantity that is borrowed from previous well-established research, such as using ``trust in automation''  \cite{anik2021data, colley2021effects, kaur2020interpreting} or ``general trust in technology'' \cite{guo2022building, kunkel2019let}. 
\pq{To further construct ``trust'' as a quantitative measurement, one needs to examine how existing work has conceptualized ``trust'' in both social sciences context as well as XAI and technical context \cite{hoffman2020primer}.}
The application-based quantity depends on the application goal, for instance in a chess game \cite{das2020leveraging}, the measurement is the human winning percentage with the help of model explanations (Human-AI \revise{collaboration}). 

\begin{figure*}
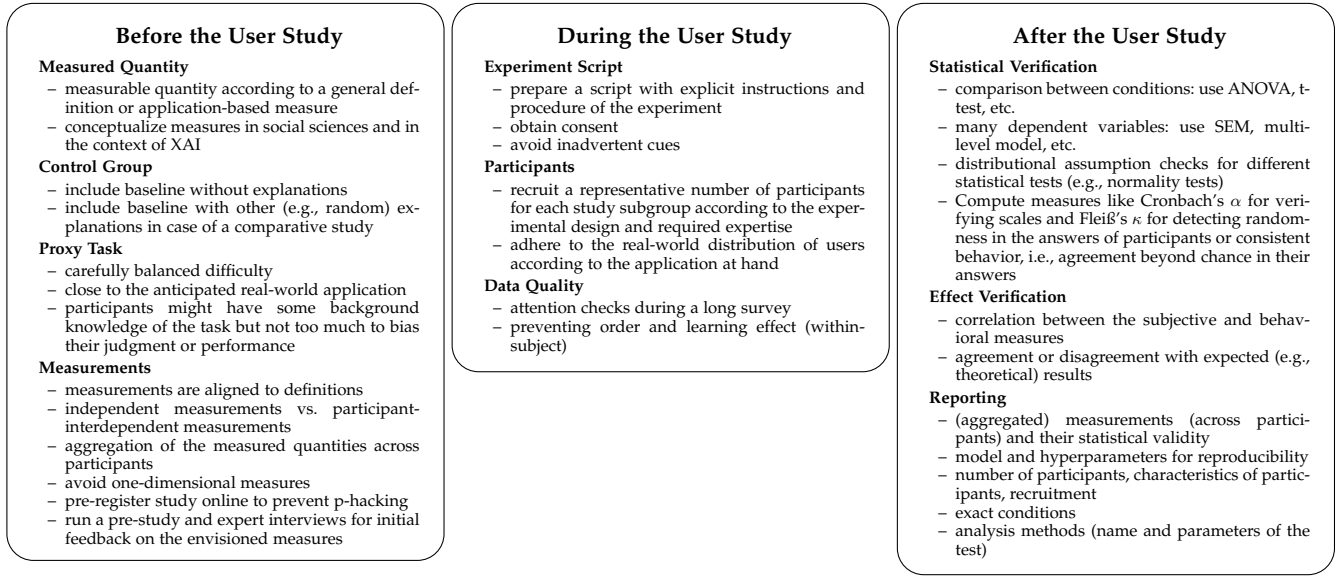

\setlength{\belowcaptionskip}{-5mm}
\centering
\begin{minipage}{\linewidth}
\begin{minipage}[t]{0.32\linewidth}
\begin{mdframed}[roundcorner=10pt]
\scalebox{0.7}{
\parbox{1.42\textwidth}{
\begin{center}
\large{\textbf{Before the User Study}}
\end{center}
\setlength{\parindent}{12pt}
\renewcommand\labelitemi{--}

\noindent \small{\textbf{Measured Quantity}}

\begin{itemize}[leftmargin=0.5cm]
    \item measurable quantity according to a general definition or application-based measure
    \item \pq{conceptualize measures in social sciences and in the context of XAI}
\end{itemize}

\noindent \small{\textbf{Control Group}} 
\begin{itemize}[leftmargin=0.5cm]
    \item include baseline without explanations
    \item include baseline with other (e.g., random) explanations in case of a comparative study
\end{itemize}
\noindent \small{\textbf{Proxy Task}} 
\begin{itemize}[leftmargin=0.5cm]
    \item carefully balanced difficulty
    \item close to the anticipated real-world application
    \item \pq{participants might have some background knowledge of the task but not too much to bias their judgment or performance}
\end{itemize}

\noindent  \small{\textbf{Measurements}}

\begin{itemize}[leftmargin=0.5cm]
    \item measurements are aligned to definitions
    \item independent measurements vs. participant-interdependent measurements
    \item aggregation of the measured quantities across participants
    \item avoid one-dimensional measures
    \item pre-register study online to prevent p-hacking
    \item run a pre-study and expert interviews for initial feedback on the envisioned measures
\end{itemize}}}


\end{mdframed}
\end{minipage} 
\begin{minipage}[t]{0.32\linewidth}
\begin{mdframed}[roundcorner=10pt]
\scalebox{0.7}{
\parbox{1.42\textwidth}{
\begin{center}
\large{\textbf{During the User Study}}
\end{center}
\setlength{\parindent}{12pt}
\renewcommand\labelitemi{--}

\noindent \small{\textbf{Experiment Script}} 
\begin{itemize}[leftmargin=0.5cm]
    \item prepare a script with explicit instructions and procedure of the experiment 
    \item obtain consent 
    \item avoid inadvertent cues
\end{itemize}

\noindent \small{\textbf{Participants}} 
\begin{itemize}[leftmargin=0.5cm]
    \item recruit a representative number of participants for each study subgroup according to the experimental design and required expertise 
    \item adhere to the real-world distribution of users according to the application at hand
\end{itemize}

\noindent \small{\textbf{Data Quality}}
\begin{itemize}[leftmargin=0.5cm]
\item attention checks during a long survey
\item preventing order and learning effect (within-subject)
\end{itemize}}}
\end{mdframed}
\end{minipage}
\begin{minipage}[t]{0.32\linewidth}
\begin{mdframed}[roundcorner=10pt]
\scalebox{0.7}{
\parbox{1.42\textwidth}{
\begin{center}
\large{\textbf{After the User Study}}
\end{center}

\setlength{\parindent}{12pt}
\renewcommand\labelitemi{--}


\noindent \small{\textbf{Statistical Verification}}
\begin{itemize}[leftmargin=0.5cm]
    \item comparison between conditions: use ANOVA, t-test, etc.
    \item many dependent variables: use SEM, multi-level model, etc.
    \item distributional assumption checks for different statistical tests  (e.g., normality tests)
     \item Compute measures like Cronbach's $\alpha$ for verifying scales and Fleiß's $\kappa$ for detecting randomness in the answers of participants or consistent behavior, i.e., agreement beyond chance in their answers
\end{itemize}

\noindent \small{\textbf{Effect Verification}}
\begin{itemize}[leftmargin=0.5cm]
    \item correlation between the subjective and behavioral measures
    \item agreement or disagreement with expected (e.g., theoretical) results
\end{itemize}

\noindent \small{\textbf{Reporting}}
\begin{itemize}[leftmargin=0.5cm]
\item (aggregated) measurements (across participants) and their statistical validity
\item model and hyperparameters for reproducibility
\item number of participants, characteristics of participants, recruitment 
\item  exact conditions
\item  analysis methods (name and parameters of the test)
\end{itemize}}}
\end{mdframed}
\end{minipage}
\end{minipage}
\caption{Summary cards of the guidelines extracted from past XAI user studies}
\end{figure*}

From \Cref{tab:findings}, we can see that previous works have frequently struggled to prove the effectiveness of XAI even with respect to a control group that is without explanation. When only different explanation techniques are considered, there will always be one winner explanation, but the overall benefit will remain undisclosed (see examples in Appendix~\ref{sec:appfindings}). Therefore, it is important to compare with a baseline without explanations to rigorously show the strength of XAI. When a comparative design is explicitly desired, baselines such as random explanations \cite{ehsan2019automated, ghorbani2019towards, schuff2022human}).

When deploying a proxy task, its difficulty should be gauged and monitored carefully. In the past, the forward simulation task has been criticized as being unrealistically complex for domains such as computer vision \cite{alqaraawi2020evaluating}. Thus, other proxy tasks such as feature importance queries \cite{sixt2022do} or manipulatability checks \cite{arora2022explain, ross2021evaluating} were proposed. Another important point is to choose a proxy task that is simplified, but features many characteristics of the application in mind \cite{doshi2017towards}. Notably, the proxy task should be designed close to the final anticipated application, as even slight differences in the tasks may void the validity of the findings on the proxy tasks in the real world \cite{buccinca2020proxy}.

The measurement is often dependent on the definition of the measured quantity. For instance, in \cite{chandrasekaran2018explanations}, the objective understanding is measured as failure prediction (the accuracy of user prediction when the model prediction is wrong). For subjective measurements such as subjective understanding or trust, one-dimensional measures (i.e., simply rating one question such as \textit{``Do you trust the model explanation?''}) have the drawback that they cannot completely reflect different constructs of measured quantities \cite{ooge2022explaining}. Moreover, subjective questions and behavioral measurements often appear to be weakly correlated. For example, the users state that they trust model but they do not really follow the model suggestions \cite{schaffer2019can}. Similar findings have been made with respect to objective and subjective understanding \cite{chromik2021think, hase2020evaluating, wang2021explanations}. To overcome this limitation, both self-reported and observed measures shall be used in parallel. 

Besides the measures introduced in \Cref{sec:measures}, there are several psychological constructs that can be deployed to evaluate multiple facets of the interaction between humans and XAI. For instance, the \textit{subjective task value} in the expectancy-value framework is often used to analyze subjective motivation to take any actions \cite{eccles1983expectancies}, which is not thoroughly studied in the XAI experience yet. The subjective task value consists of intrinsic value (enjoyment), attainment value (importance for one's self), utility value (usefulness), and cost (the amount of effort or time needed) \cite{eccles1983expectancies,hulleman2017making}. A good explanation interface should be positively correlated with the subjective task value, consequently boosting one's interest and motivation to use the model explanation. With regard to the cost of using model explanations, cognitive load is popularly measured in the current literature with conventional Likert scales \cite{hart1988development,paas1992training}. Cognitive load researchers study the validity of different visual appearances in rating scales beyond numerical Likert scales, i.e., pictorial scales such as emoticons (faces with different emotions), or embodied pictures of different weights \cite{ouwehand2021measuring}. Their results demonstrate that numerical scales are more proper in complex tasks while pictorial scales are for simple ones.

Pre-registration using online platforms such as AsPredicted\footnote{\url{https://aspredicted.org}} has become a common practice in recent years \cite{Simmons2021preregisttration}. In this process, researchers submit a document detailing their planned study online before initiating the data collection. Among other details, the pre-registration includes the measured variables and hypotheses, data exclusion criteria, and the number of samples that will be collected. An exhaustive pre-registration can provide evidence against the findings being a result of selective reporting or p-hacking \cite{simonsohn2014p} and thus strengthen the credibility of a study. Expert interviews and pre-studies following a think-aloud protocol \cite{ericsson1984protocol}, e.g., in the references \cite{ross2021evaluating, zhang2022towards}, are often mentioned as helpful tools to develop the explanation system and the study design and gain first qualitative insights or complement the qualitative analysis \cite{buccinca2020proxy, wang2022interpretableideation}.

\pq{When preparing for a user study, it is important to plan for explicit steps and to have a backup plan for different situations. Before participants arrive, it is helpful to provide them with information such as where the researchers will meet with them, what they need to bring, and how they can prepare for the study. 
If conducting the experiment in person, send participants a reminder the day before and provide them with your contact in case they cannot find the experiment site or they need to cancel the experiment session. Once participants arrive, make sure the researchers have a plan that covers all stages of the experiment. 
The protocol should cover small details (e.g., where participants should leave their backpacks, water bottles, and lunch boxes) and plans for unexpected situations (e.g., uncooperative participants and multifunctional systems). 
How to obtain participants' consent should be an important part of the procedure. Additional procedure is required for obtaining consent when working with vulnerable populations (e.g., children and pregnant women), in which case alternative consent procedures might take place. 
Another benefit of pre-designing the experiment script is to fine-tune the language to avoid inadvertent cues. Researchers can unintentionally pass on their expectations to participants through verbal and nonverbal behavior, which might result in participants' skewed performance towards the researchers' desire \cite{hoffman2020primer}.
To ensure a sound experiment procedure and to protect the integrity of the data, it is worthwhile to put in much effort to design a detailed experiment script.
}
\vspace{-5pt}
\paragraph*{During the User Study} 
A sufficient number of participants is the prerequisite of a solid user study analysis. To get a rough estimate of common sample sizes, we refer the reader to the participant statistics in \Cref{fig:user statistics} where we analyze the subject numbers in different experimental designs. For instance, around 350 users without any specific expertise are averagely recruited in between-subject experiments. However, we would like to underline that the required number of participants is highly specific to the study design and should be determined individually, for instance by conducting a statistical power analysis~\cite{cohen2013statistical}. Additionally, recruited participants should have the same knowledge background as the end users that applications are designed for. For instance, when evaluating an interface explaining loan approval decisions to bank customers, it is not proper to include only students whose major is computer science, since they may have prior knowledge of how model explanations work. Note that the design of an AI application requires different audiences across the project cycle, thus model explanations need to evolve as well \cite{dhanorkar2021needs}.

To uphold high-quality standards of the collected data, attention or manipulation checks are essential to filter out careless feedback. This particularly applies to long surveys or online surveys with lay users. Kung et al.~\cite{kung2018attention} justify the use of these checks without compromising scale validity. In within-subject experiments, a random order of conditions is necessary to avoid order effect \cite{panigutti2022understanding}. Participants can learn knowledge of data or examples shown in the previous conditions, and Tsai et al.~\cite{tsai2021exploring} choose to use a Latin square design to avoid the learning effect. 
\vspace{-5pt}
\paragraph*{After the User Study} After the data collection, statistical tests are run to find significant effects. The applicable tests used are determined by experimental designs and the form and distribution of the data. Generally, ANOVA tests and T-test are usually used when comparing distributions between different conditions. Structural Equation Models (SEM) or multi-level models are used for mediation analysis. More details of statistic tools can be found in \Cref{sec:analysis}. Distributional assumption checks should be applied. When Likert-type data is collected as in most of the questionnaires, non-parametric tests such as paired Wilcoxon signed-rank test, or Kruskal-Wallis H test for multiple groups can be used to avoid normality assumptions.
\vspace{-1pt}

If multiple measures are aggregated into a single instrument, it is important to assess the validity of this aggregation with reliability measures such as the tau-equivalent reliability (also known as Cronbach's $\alpha$). For example, if objective and subjective measures of a quantity, such as understanding are combined, it is necessary to verify that there is sufficient agreement.
If multiple items (e.g., data samples or visualizations) are rated by several subjects, statistics such as Cohan's $\kappa$ as Fleiß's $\kappa$ for more than two raters \cite{fleiss1971measuring} can be used to assess agreement beyond chance between these raters and serve as an indication for the reliability of the ratings.
\vspace{-1pt}

In the final writing phase, it is essential to report sufficient details that allow readers to estimate the explanatory power of the study. On the level of participants, this should include the total number of participants and how many are assigned to each treatment group, their recruitment, consent and incentivization, and the exact treatment conditions they are subjected to.
Furthermore, some descriptive statistics of the collected data can help readers assess the characteristics of the adequacy of the statistical tools used. Regarding the analysis, we found it important to mention how the underlying assumptions of the statistical tests used were checked and to mention the exact variant of the test used (e.g., stating ``a two-way ANOVA with the independent variables X and Y'' is used instead of just mentioning that ANOVA-test is used).

\vspace{-7pt}
\section{Future Research Directions}
\label{sec:summary}
\vspace{-4pt}
Our survey of recent and ongoing XAI research also helps us identify research gaps and distill a few directions for future investigations. In this section, we highlight these directions and summarize our findings.
\vspace{-9pt}
\subsection{Towards Increasingly User-Centered XAI}
\label{sec:user-models}
We advocate that user-centered methods should be used not only to assess XAI solutions (e.g., through user studies) but also to design them (e.g., through user-centered design). By explicitly modeling and involving users in the design phase and not just in a post-hoc manner during the evaluation phase, we expect the development of XAI solutions that better respond to user needs. As discussed in~\cite{riedl2019human}, there are two aspects of human-centered AI: (1) AI systems that understand humans with a sociocultural background and (2) AI systems that help humans understand them. The former point can guide the design of AI systems. In this section, we discuss XAI research that leverages this insight.

The process of explaining a machine's decisions to human users can be viewed as a teaching-learning process where the XAI system is the teacher and the human users are the students. From a user-centered perspective, the problem of designing effective teaching methods to enhance the student's (i.e., user's) learning outcomes is essential to human-centered XAI algorithms. To leverage the ability of humans and address unique user's needs, it is important to review studies and findings from psychology and education. These studies provide insights into how humans perceive other intelligent agents (humans or artificial agents) and how they utilize limited information to infer and generalize. Understanding how humans think and learn will help XAI developers build and design systems that are not only informative but also user-friendly to people with different backgrounds.
In this section, we discuss three pedagogical frameworks, namely (1) the expectancy-value motivation theory, (2) the theory of mind, and (3) hybrid teaching, to shed light on incorporating such methods in computational approaches. Inspired by existing work in pedagogy and XAI, we provide implications for designing future transparent AI systems and human-centered evaluations. 

\vspace{-3pt}
\paragraph*{Expectancy-value Motivation Theory}
Human interaction with XAI interfaces can be viewed as an activity where humans learn about the model's inner workings through explanations and then achieve an understanding of the models. 
The question of how to enhance the efficiency and the outcome of this human learning process is of high importance \cite{lage2019exploring}. This research problem is widely considered in educational psychology through the lens of expectancy-value motivation theory. 
For instance, Hulleman et al.~\cite{hulleman2017making} propose to utilize \textit{interventions} to increase the perception of usefulness (utility value) to subsequently increase motivation and final performance. Intervention here refers to identifying the relevance of model explanations to the user's own situation, which can be a prompt question while working with the interface. Moreover, when utilizing model explanations in human-AI collaboration, explanations can be seen as a type of ``scaffolding'' (prompt during a task) proposed in a conceptual framework in education.


\vspace{-3pt}
\paragraph*{Theory of Mind}
When interacting with XAI systems, humans form mental models of the machine learning algorithms that reflect their belief of how the algorithms work. The formation of these mental models comes from observing explanations or examples given to the human, who often subconsciously applies the observations in a few examples to the broader understanding of the whole machine learning system. This incredible ability to infer, rationalize, and summarize other intelligent agent's decisions is known as the Theory of Mind (ToM) in psychology.
Based on this theory, the Bayesian Theory of Mind (BToM) provides a probabilistic framework to predict inferences that people make about mental states underlying other agents' actions. 
Recent work, at the intersection of XAI and robotics, indicates that humans also attribute ToM to artificial agents that they observe or interact with.
Guided by these user-centered results, several works at the intersection of XAI and robotics have utilized BToM to create a simulated user, and then use it to generate helpful explanations.


\vspace{-4pt}
\paragraph*{Hybrid Teaching}
Teaching strategies for the human-to-human setting have been widely studied and many categorizations exist.
One way of categorizing these strategies is through the following three concepts: (1) direct teaching, (2) indirect teaching, and (3) hybrid teaching.
\textit{Direct teaching} utilizes direct instructions that are teacher-centered, involve clear teaching objectives, and are consistent with classroom organizations. In XAI applications, direct teaching methods generate explanations by selecting representative examples of an agent's decisions to convey the patterns in its policy. 
In contrast, \textit{indirect teaching} is student-centered and encourages independent learning. In the XAI perspective, methods utilizing indirect teaching provide users with tools to actively and independently explore an AI system.
Technically, direct teaching focuses on providing guidance (using a computational approach) to assist users in building an understanding of a machine, whereas indirect teaching (often through a user interface) enables users to address individual learning preferences and mitigate individual confusion about the AI.
To leverage the advantages of the two teaching strategies, \textit{hybrid teaching} has been widely used in human-to-human teaching with an emphasis on interactivity.
Recent work ~\cite{qian2022evaluating} indicates that hybrid teaching reduces the amount of time for a user to understand an agent's policy compared to direct and indirect teaching, and is more subjectively preferred by the participants.  
Building on this, future XAI systems can consider using hybrid teaching methods that $(i)$ generate direct instructions to provide guidance to user's understanding of an AI system; and $(ii)$ provide methods to allow users to interact with the agent. 

\vspace{-12pt}
\revise{\paragraph*{Explanations through Large Language Models (LLMs)} The recent rise of Large Language Models \cite{radford2019language, chatgpt2022optimizing} naturally opens up new research directions. There is a growing interest in leveraging their unprecedented capabilities \cite{bubeck2023sparks} to offer explanations for model decisions \cite{zhou2020towards, wiegreffe2022reframing}. Through their natural language interface, LLMs offer the possibility to build interactive explainers \cite{wang2023chatcad}. Intriguingly, textual explanations can also be used as subsequent inputs to LLMs which may help to solve subsequent problems and result in superior performance \cite{rajani2019explain}. This technique, referred to as chain-of-thought reasoning \cite{wei2022chain}, opens up an interesting research territory combining interpretability and performance considerations.}

\vspace{-5pt}
\subsection{Open Research Problems}
\subsubsection{Automatic vs. human-subject evaluations} 
With automatic evaluations, we refer to evaluation methods that do not require human subjects, which corresponds to the functionally-grounded metrics discussed in \cite{doshi2017towards,nauta2023anecdotal}. These metrics aim to test desiderata around the ``faithfulness''/``fidelity''/ ``truthfulness'' of model explanations \cite{nauta2023anecdotal,tomsett2020sanity,alvarez2018towards}. Faithfulness of explanations is defined as that explanations are indicative of true important features in the input \cite{alvarez2018towards}. The automatic evaluations aim at capturing general objectivity which is independent from downstream tasks, while human evaluations are contextualized with specific use cases. Generally speaking, automatic evaluations and human evaluations tackle different research challenges: the former objectively examines how truly explanations reflect models and the latter one measures how humans perceive models through explanations (although there existing algorithms for automated evaluation designed to align with human evaluations, which we will discuss later). 
All explanations used in human-subject experiments should have satisfying performance in automatic evaluations, i.e., the explanations should be able to faithfully unbox the model.
This verification step is essential to guarantee the validity of the empirical user study and to ensure that users are not tricked by unfaithful explanations. However, in most current human-subject experiments, the functional faithfulness of explanations is not thoroughly verified beforehand. Using unfaithful explanations could lead to the problem that only the placebo effect of explanations is measured. Ideally, a good explanation should be faithful to the model as well as understandable by users. 

\vspace{-6pt}
\subsubsection{\revise{Identifying and handling confounders}} \revise{Existing research underscores the vulnerability of model explanation studies to significant confounding effects. For instance, Papenmeier et al.~\cite{papenmeier2019model} reveal that user trust can be more influenced by model accuracy than the faithfulness of the explanation itself. Similarly, Yin et al.~\cite{yin2019understanding} demonstrate that the accuracy score perceived by users and the one shown to users contribute to trust formation.} 

\revise{A different problem is that good explanations also reveal weaknesses of the model. However, when seeing unexpected explanations, users may express their negative feelings about the model through negative ratings of the explanations. Therefore, good model explanations should help users \textit{calibrate} their trust \cite{bussone2015role, rong2022user}, i.e., trust the model's decision when it is correct but distrust it otherwise. There is a disagreement on how to handle such cases: When evaluating model fairness, several works \cite{harrison2020empirical, schoeffer2022there, rader2018explanations, anik2021data, grgic2018human} reckon the increase in perceived fairness as positive, while Dodge et al.~\cite{dodge2019explaining} define the decrease as positive.
Other factors, such as the temporal occurrence of model errors (Nourani et al.~\cite{nourani2021anchoring}), and the dimensions of models (Ross et al.~\cite{ross2021evaluating}, Poursabzi et al.~\cite{poursabzi2021manipulating}), also come into play.}

\revise{In summary, these confounding elements suggest that users might be led to put more trust in oversimplified, deceptive, or simply unfaithful explanations. To mitigate this, we recommend meticulous analysis, control and reporting of potential confounders, such as explanation faithfulness and model accuracy, across various test conditions. More advanced measures have been suggested as well. For instance, Schoeffer and
Kuehl's~\cite {schoeffer2021appropriate} propose \textit{appropriate fairness perceptions}, which measures whether people increase or decrease their fairness perceptions depending on the algorithmic fairness of the underlying model.
Nevertheless, the thorough investigation of confounding factors remains a challenge. Calibrated measures that are less prone to confounding can be a valuable step forward.}


\vspace{-12pt}
\revise{\subsubsection{Mitigating personal biases for XAI}} 
\revise{Most XAI techniques and corresponding designed user studies provide \textit{one-size-fits-all} solutions. 
Individual bias, rooted in a user's mental framework, influences the user's perception of a model. It should be considered in XAI design, development, and evaluation procedures. Several studies that aim to explain reinforcement learning policies utilize cognitive science theories to create a model of the human user~\cite{baker2011bayesian,huang2019enabling, lage2019exploring, qian2022evaluating}. 
They then generate explanations based on this human model and verify the benefits of tailoring explanations for individual user models. Within the scope of XAI, \cite{yang2022psychological,yang2021mitigating} utilize a Bayesian Teaching framework to capture human perception of model explanations. In user studies, depending on cultural and educational background, participants may likely give different feedback~\cite{springer2019progressive}. This kind of personal bias can be mitigated by deploying a large sample size and recruiting participants who are representative of the target audience. We advocate that personal biases should be taken into account in the realm of XAI development.}

\vspace{-7pt}
\revise{\subsubsection{Human-in-the-Loop and sequential explanations} In several relevant cases, such as online recommendation systems, users are not only confronted with an explanation once but instead view decisions and potential explanations repeatedly. Recent work in this domain \cite{chromik2021think} has shown that the order of decisions and explanations may indeed have an effect on user perception and understanding. The AI model may continue to shape the user's mental model over time. The differences between the single-use and the sequential setting still remain to be thoroughly investigated.}

\subsubsection{Proxy tasks should be close to real-world tasks} When using proxy tasks to evaluate models, for instance, to measure subjective understanding, there is a great choice of tasks present in the literature. 
A good proxy task should have the following features: (1) it has close real-world connections \cite{doshi2017towards}; (2) users or participants have some background knowledge of the task but not too much to affect their judgment or performance during the task; (3) the task is not too complicated to implement or there exists an existing implementation but was used for different purposes (i.e., not used for XAI); and (4) it has connections to existing work.
Yet, the link between evaluations through different proxy tasks and real-world applications has not been made very explicit to date. Buçinca et al.~\cite{buccinca2020proxy} show that the outcomes of proxy evaluations can be different from a real-world task. More specifically, the widely accepted proxy tasks, where users are asked to build the mental models of the AI, may not predict the performance in actual decision-making tasks, where users make use of the explanations to assist in making decisions. The results show that users trust different explanations in the proxy task and the actual decision-making task. 
Therefore, we argue that further research is required to uncover the links between current proxy tasks and on-task performance or to devise new proxy tasks with a verified connection to actual tasks. 
\vspace{-7pt}
\subsubsection{Simulated evaluation as a cost-efficient solution} As human-subject experiments are costly to conduct, Chen et al.~\cite{chen2022use} propose a simulated evaluation framework (\texttt{SimEvals}) to select potential explanations for user studies by measuring the predictive information provided by explanations. Concretely, the authors consider three use cases where model explanations are deployed: forward simulation, counterfactual reasoning, and data debugging. Human performance is measured for these three tasks with different explanations. If there is a significant gap in settings of using two types of explanations,
the simulated evaluation can also observe such a gap under the same task settings as well. Meanwhile, first attempts to simulate human textual responses in a given context using large language models show that models can provide surprisingly anthropomorphic answers \cite{aher2022using}. Undoubtedly and also affirmed by Chen et al.~\cite{chen2022use}, it is not yet realistic to replace human evaluation with the simulated framework as other factors e.g., cognitive biases can affect human decisions. To better simulate human evaluations, more effort should be directed towards modeling human cognitive processes. Concurrently and with appropriate caveats, XAI researchers should also leverage existing and approximate models of human cognition to enable rapid prototyping and assessment of explanations. Section~\ref{sec:user-models} discusses several candidate human cognition models and highlights recent XAI works \cite{lage2019exploring, qian2022evaluating} that utilize this ``Oz-of-Wizard'' paradigm. 

\vspace{-8pt}
\section{Conclusion}
In recent years, there has been a proliferation of XAI research in both academia and industry. Explainability is a human-centric property \cite{liao2021human} and therefore XAI should be preferably studied by taking humans' feedback into account. In this work, we investigated recent user studies for XAI techniques through a principled literature review. Based on our review, we found out that the effectiveness of XAI in users' interaction with ML models was not consistent across different applications, thus suggesting that there is a strong need for more transparent and comparable human-based evaluations in XAI. Furthermore, relevant disciplines, such as cognitive psychology and social sciences in general, should become an integral part of XAI research. 

We comprehensively analyzed the design patterns and findings from previous works. Based on best-practice approaches and measured quantities, we propose a general guideline for human-centered user studies and several future research directions for XAI researchers and practitioners. Thereby, this work represents a starting point for more transparent and human-centered XAI research.


\bibliographystyle{IEEEtran}
\bibliography{sample-base}

%



%




\vspace{-17mm}
\begin{IEEEbiography}[{\includegraphics[width=1in,height=1.2in,clip,keepaspectratio]{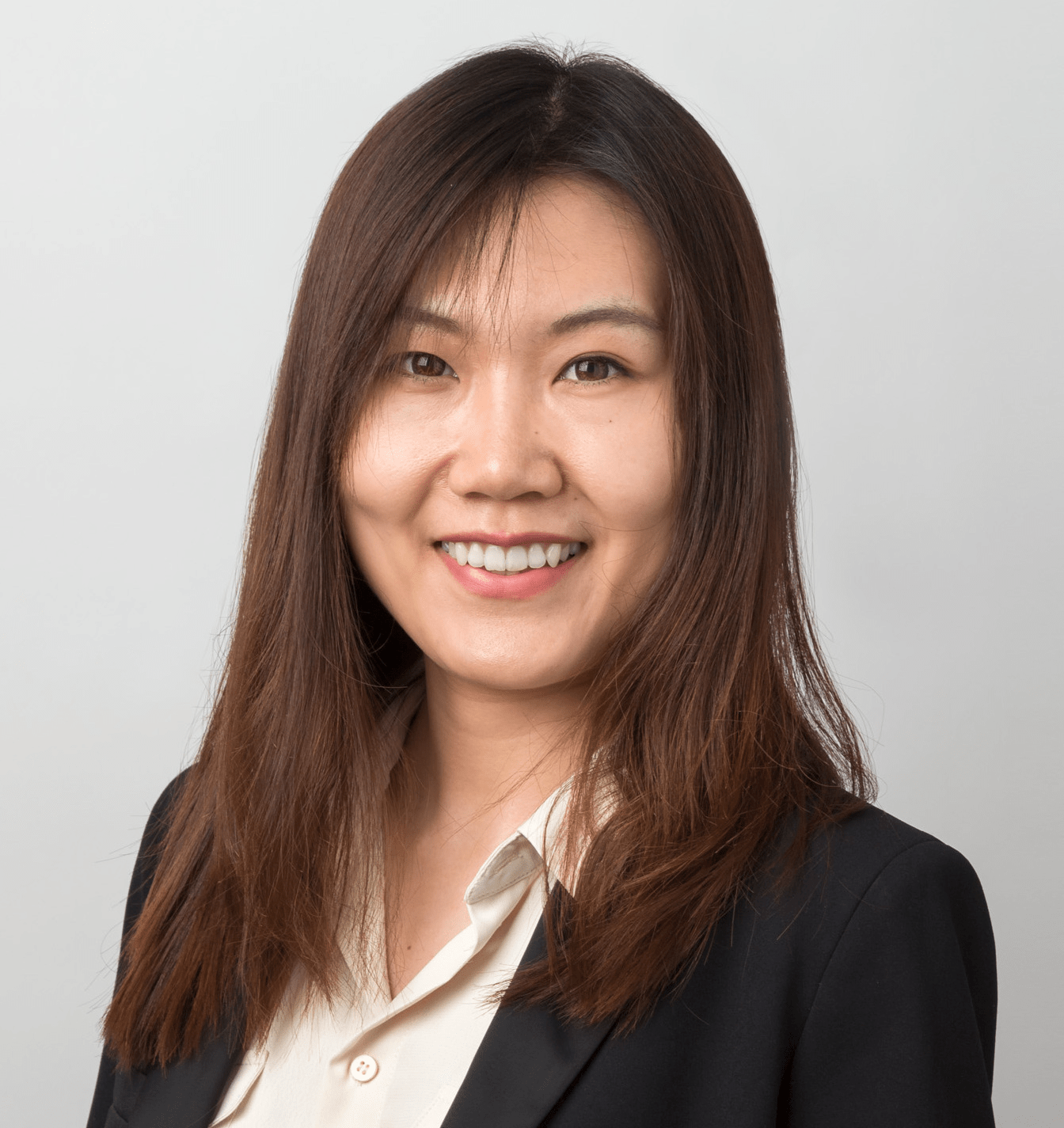}}]{Yao Rong} received her M.Sc. degree in electrical and computer engineering from the Technical University of Munich, Germany, in 2019. 
She is currently pursuing her doctoral degree with the Human-Centered Technologies for Learning Group at the Technical University of Munich. From September 2022 to February 2023, she served as a visiting scholar at the DATA Lab at Rice University. Her research interests lie in human-centered AI,  explainable AI, and human-AI interaction technologies. 
\end{IEEEbiography}
\vspace{-18mm}
\begin{IEEEbiography}[{\includegraphics[width=1in,height=1.2in,clip,keepaspectratio]{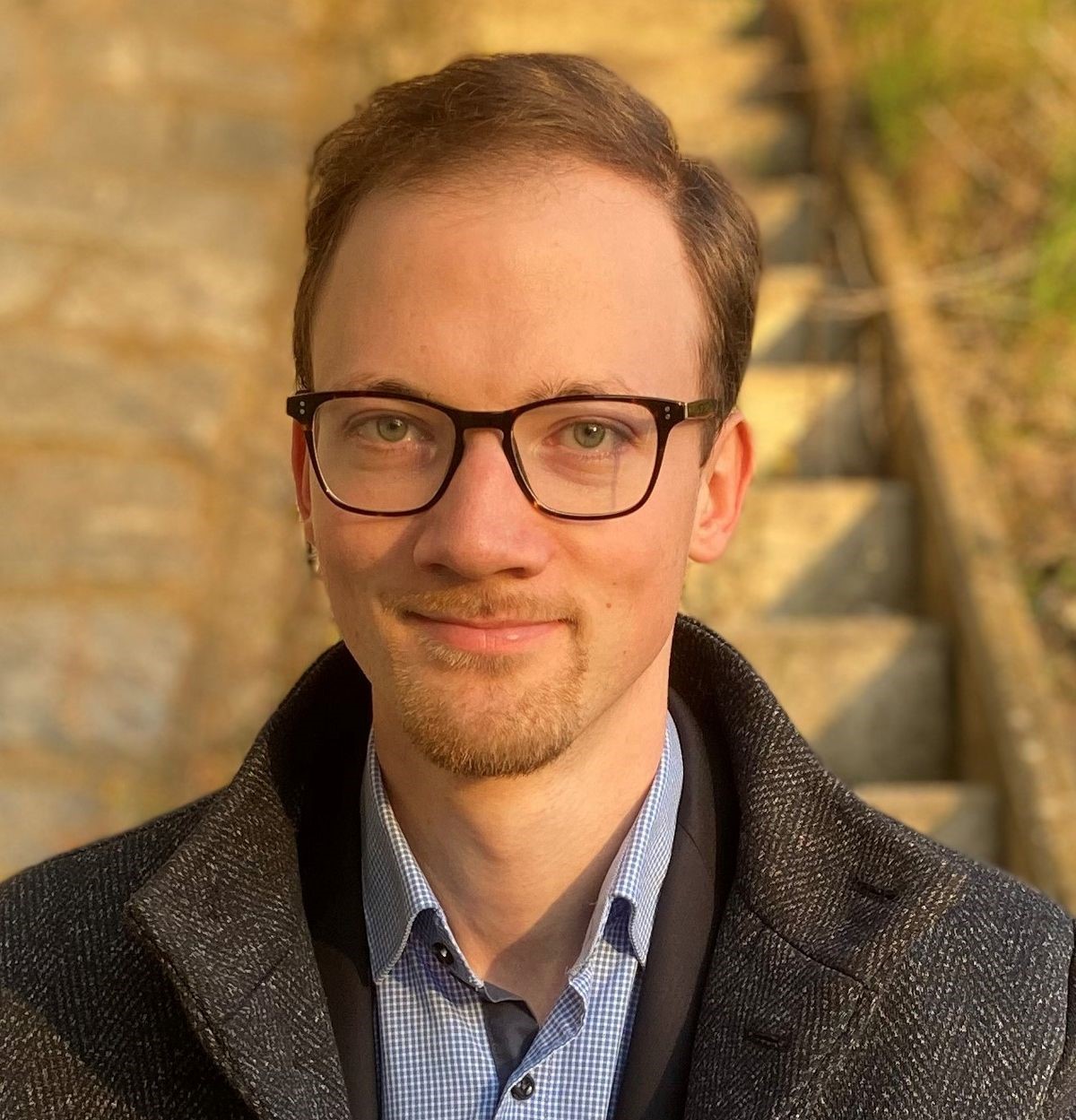}}]{Tobias Leemann} received his M.Sc. degree from the University of Erlangen-Nuremberg, Germany in 2020 and is currently pursuing the Ph.D. degree at the University of Tübingen, Germany where his research is focused on trustworthy machine learning. Specifically, his research interests include the quality assessment of interpretability techniques and the intersections of interpretability, fairness and privacy.
\end{IEEEbiography}
\vspace{-16mm}
\begin{IEEEbiography}[{\includegraphics[width=1in,height=1.25in,clip,keepaspectratio]{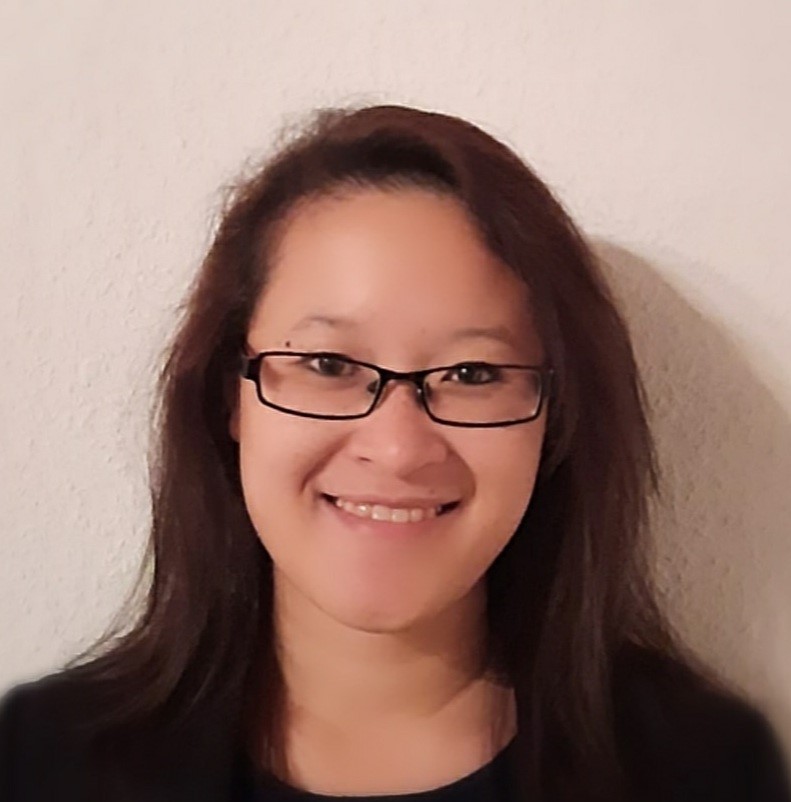}}]{Thai-Trang Nguyen}
In 2021, Thai-Trang Nguyen graduated with a B.Sc. degree in computer science at the University of T\"ubingen, Germany. She is currently pursuing her M.Sc. degree at the same university. Furthermore, she served as a research assistant at the Human-Computer Interaction group from October 2019 to October 2022.
\end{IEEEbiography}
\vspace{-16mm}
\begin{IEEEbiography}[{\includegraphics[width=1in,height=1.25in,clip,keepaspectratio]{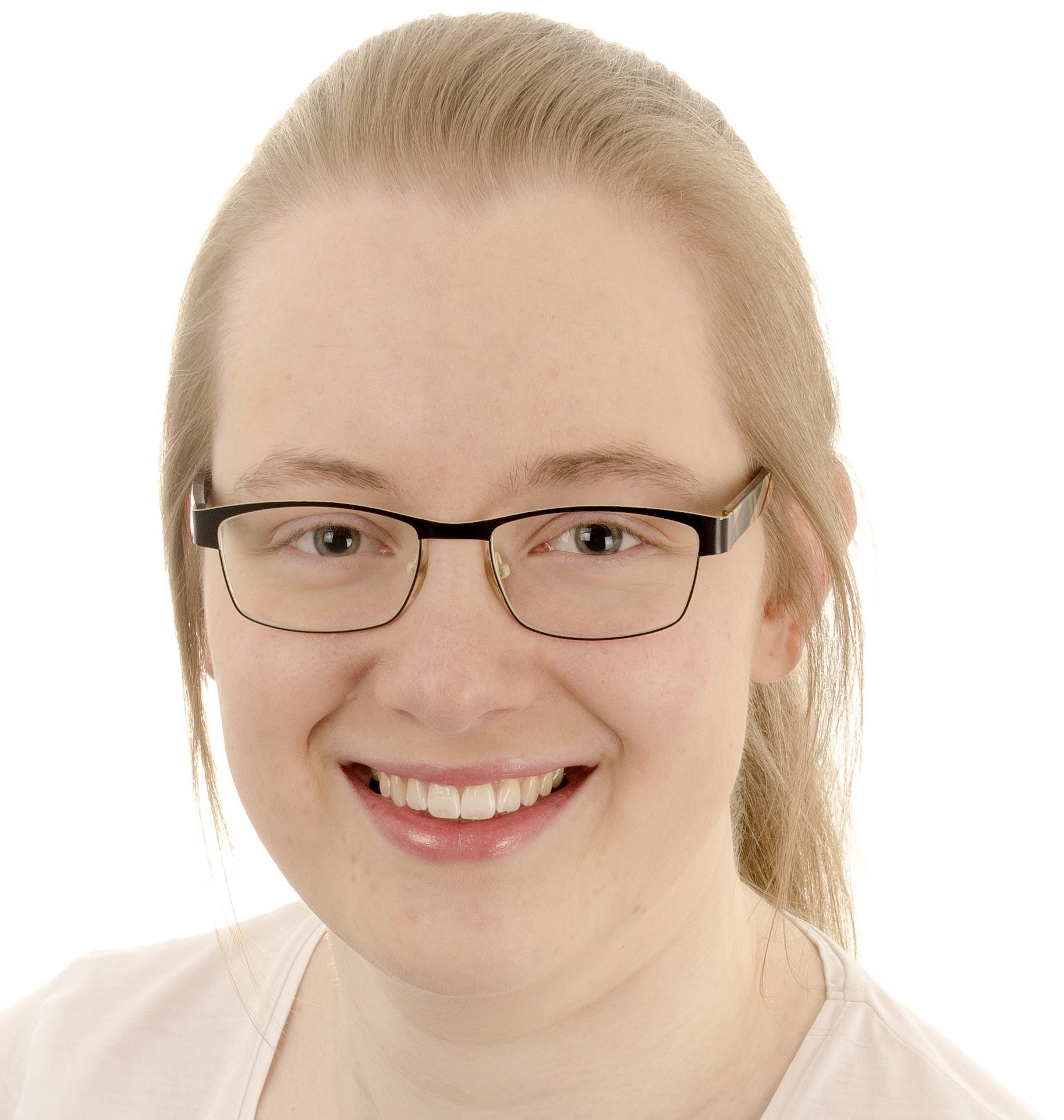}}]{Lisa Fiedler}
Lisa Fiedler is currently pursuing her B.Sc. degree in media informatics at the University of Tübingen, Germany. Additionally, she works as a student assistant for the Human-Computer Interaction Group at the University of T\"ubingen.
\end{IEEEbiography}
\vspace{-18mm}
\begin{IEEEbiography}[{\includegraphics[width=1in,height=1.25in,clip,keepaspectratio]{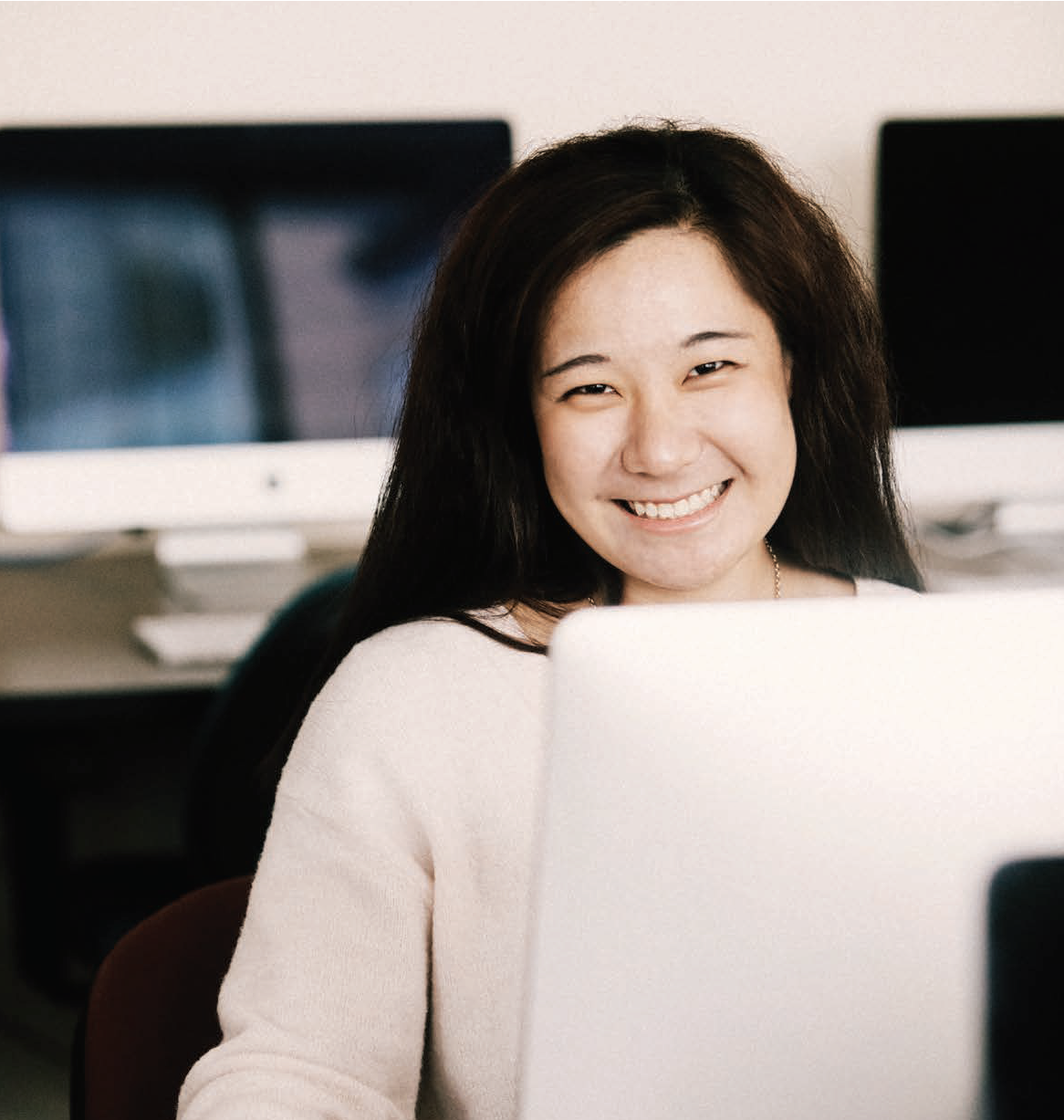}}]{Peizhu Qian}
is a PhD candidate in Computer Science at Rice University, USA working with Dr. Vaibhav Unhelkar on problems in human-robot interaction, robot transparency, and explainable AI. Her research interest lies in building a mutual understanding between a robot and its human collaborators. Her work applies psychology theories to computational frameworks, enabling robots to communicate their objectives. 

\end{IEEEbiography}
\vspace{-18mm}
\begin{IEEEbiography}[{\includegraphics[width=1in,height=1.25in,clip,keepaspectratio]{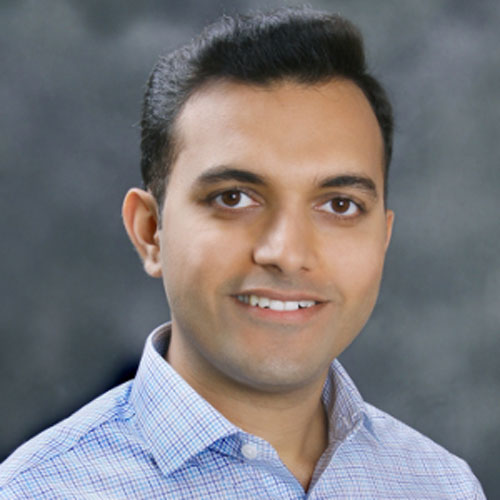}}]{Vaibhav Unhelkar}
is an Assistant Professor of Computer Science at Rice University, USA where he leads a research group in the emerging area of Human-Centered AI and Robotics. 
Unhelkar earned his undergraduate degree in aerospace engineering from the Indian Institute of Technology in Bombay in 2012. From the Massachusetts Institute of Technology, where he worked in the Computer Science and Artificial Intelligence Laboratory (CSAIL), he earned an M.S. in aeronautics and astronautics and a Ph.D. in autonomous systems in 2015 and 2020, respectively. 
\end{IEEEbiography}
\vspace{-12mm}
\begin{IEEEbiography}[{\includegraphics[width=1in,height=1.25in,clip,keepaspectratio]{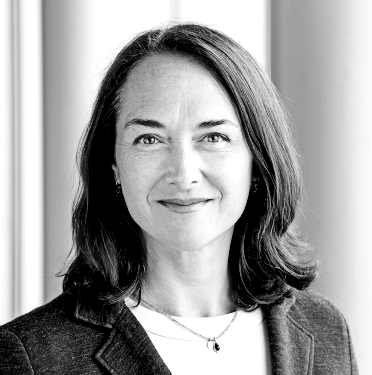}}]{Tina Seidel} holds the Friedl Schoeller Chair for Educational Psychology at the School of Social Sciences and Technology, Technical University of Munich, Germany. She received her diploma in psychology from the University of Regensburg (Germany) and Vanderbilt University Nashville (USA) in 1998. 
She received her Ph.D. with excellence in 2002 from the Leibniz Institute for Science and Mathematics Education Kiel (Germany). Her research focuses on teaching and teacher education. She has established a Teacher Research \& Training Simulation Center that conducts several research projects funded by the German Science Foundation and the German Federal Ministry of Education and Research. 
\end{IEEEbiography}
\vspace{-12mm}
\begin{IEEEbiography}[{\includegraphics[width=1in,height=1.25in,clip,keepaspectratio]{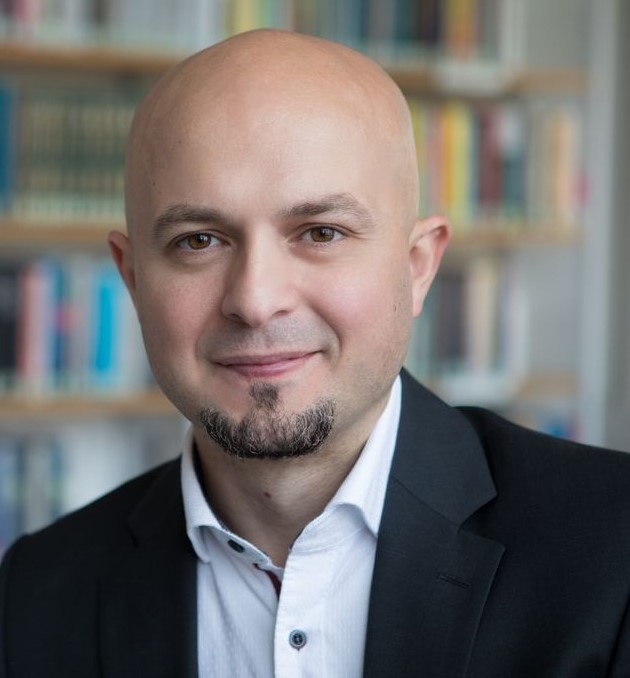}}]{Gjergji Kasneci} obtained his MSc of Computer Science and Mathematics from the University of Marburg in 2005 and his PhD from the University of Saarland - while at the Max Planck Institute - in 2009. He then worked at Microsoft Research Cambridge, the Hasso Plattner Institute, and SCHUFA Holding AG, where he served as CTO from 2017 to 2022.  
Between 2018 and 2023, he led the Data Science and Analytics Group at the University of Tübingen as an Honorary Professor. In 2023, Gjergji Kasneci was appointed Professor of Responsible Data Science at the Technical University of Munich.

\end{IEEEbiography}
\vspace{-12mm}
\begin{IEEEbiography}[{\includegraphics[width=1in,height=1.25in,clip,keepaspectratio]{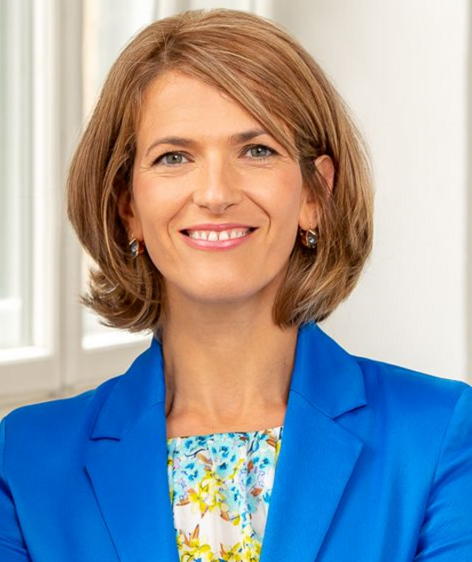}}]{Enkelejda Kasneci} is a Distinguished Professor for Human-Centered Technologies for Learning at the Technical University of Munich and Core Member of the Munich Data Science Institute. 
In 2013, she received her PhD in Computer Science from the University of Tübingen. 
and was a postdoctoral researcher and a Margarete-von-Wrangell Fellow at the University of Tübingen. 
Her research evolves around Human-Centered Technologies and AI systems that sense and infer the user's cognitive state, the level of task-related expertise, actions, and intentions based on multimodal data and provide information for media and assistive technologies in many activities of everyday life, and especially in the context of learning.
\end{IEEEbiography}




\newpage
\appendices
\section{\rev{Data-driven Bibliometric Analysis}}
\label{appendix:data-driven}
\begin{figure*}[h]
\centering
\includegraphics[width=.95\linewidth]{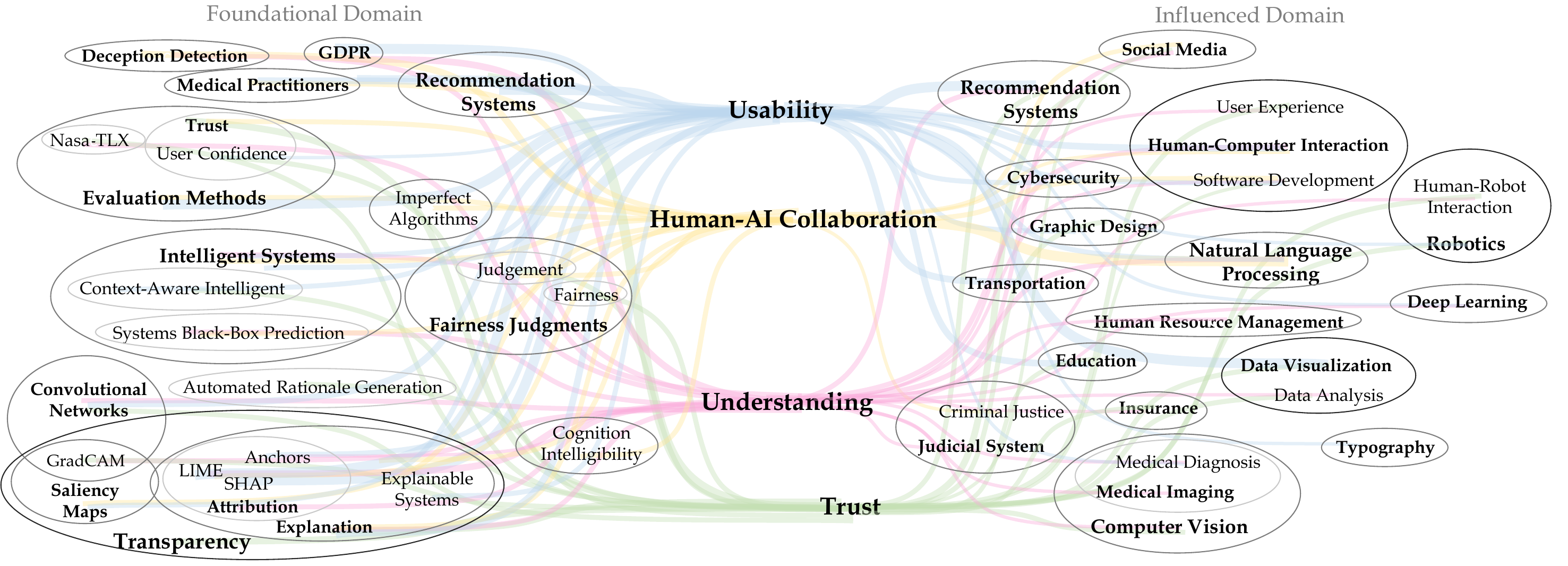}
\caption{\rev{Illustration of the \textbf{foundational} research domains (\textbf{Left})}: Each dot represents a referenced paper, whose size reflects the number of studied core papers referring to it. \rev{Illustration of \textbf{influenced} research domains (\textbf{Right})}: Each dot represents a research topic, whose size refers to the number of papers on the same topic.
For a clear depiction, only several important research domains are labeled with text. Lines are used to depict reference links, with thicker lines representing a greater number of links. Core paper categories are in blue (\textbf{Middle}). \rev{Circles are used to indicate a hierarchical structure of keywords.}}
\label{appendix:combined}
\end{figure*}

To perform a data-driven bibliometric analysis of the references and citations for all papers\footnote{In this section, the word ``references'' refers to sources contained in the references of one of the core papers while ``citations'' refers to follow-up works that reference one of the core papers}, we first collected common references from each category. As we had to deal with a large number of papers, a keyword representing the research topic was assigned to each paper. In this way, we could group the papers according to their content. Concretely, the references were extracted directly from the studied papers (in pdf format). The follow-up works that cite each core paper were retrieved from the Google Scholar platform using the Python API (``Scholarly''~\cite{cholewiak2021scholarly}). The same API was used to extract abstracts from Google Scholar for all references and citations. 
\revise{Based on the paper titles and abstracts, we utilized GPT-4 \cite{brown2020language} to tag the papers with keywords and subsequently reviewed the sensibility of these keywords manually.} We visualized papers in a 2-dimensional semantic space according to their keyword embeddings using t-SNE \cite{van2008visualizing}. 

We illustrate the research domains that are fundamental to XAI user studies in \revise{\Cref{appendix:combined} (\textbf{Left})}. Note that for presentation clarity, we only visualized works that were used as references in at least five of the core papers. Similarly to foundations in XAI user studies, we are interested in knowing who will eventually benefit from the findings of XAI user studies. \revise{\Cref{appendix:combined} (\textbf{Right})} demonstrates the ``consumers'' of the human-centered XAI core papers (i.e., research domains influenced by the core papers), with each dot representing a research topic. The size of the dots is determined by the number of citations in the set of core papers obtained from this research area. 

By studying these two aspects (i.e., foundations and impact), we grasp a clear overview of relevant topics in the research landscape of XAI user studies. More importantly, we can better spot the nascent but pertinent areas for future work such as cognition-driven analysis tools in XAI. We release raw data and code for analyses at \url{https://github.com/yaorong0921/hxai-survey}.

\subsection{Foundation of XAI User Studies}
\label{appendix:foundation}
Through analyzing references in the core papers, we provide XAI researchers with several indispensable literature sources in this field, which can inspire them when organizing their projects. In total, there are over 3000 references from all the core papers, and we pay close attention to the references which are cited at least by ten core papers (ca. 50 papers). In \Cref{tab:foundation}, we categorize these papers according to their topics.
The first group of papers is survey papers about XAI, which are thoroughly discussed in Sec.2 Related Work. For the theory of XAI, Miller et al.~\cite{miller2019explanation} propose to build XAI on social sciences such as cognitive science and psychology, while Wang et al.~\cite{wang2019designing} and Liao et al.~\cite{liao2021question} provide theoretical guidelines for designing XAI frameworks. An important class of references are XAI methods and the most popularly used ones are listed in ``XAI Methods''. As suggested by \cite{kulesza2015principles,kulesza2013too}, the explanations should be sound and complete and thus bring a positive impact on users. Another motivation for XAI is that it should assist users in building mental models of the AI systems \cite{kulesza2012tell}. Previous user studies for ML systems or for explainable interfaces that are referenced for comparisons or serve as templates of user study design. In the end, we list several general works about user trust that may go beyond the scope of XAI.
\begin{table*}[t]
\begin{tabular}{c|c}
\hline
\textbf{Topic} & \textbf{Fundamental works}  \\ \hline
 Surveys of XAI & \cite{doshi2017towards}, \cite{lipton2018mythos}, \cite{abdul2018trends}, \cite{adadi2018peeking}, \cite{gilpin2018explaining}, \cite{hoffman2019evaluating}, \cite{arrieta2020explainable} \\ \hline
 
 Theories for XAI & \begin{tabular}[c]{@{}c@{}} \cite{miller2019explanation}: social sciences, \cite{wang2019designing}: theory for XAI design, \\ \cite{liao2021question}: a question bank for XAI design \end{tabular} \\ \hline
 
 XAI Methods & \begin{tabular}[c]{@{}c@{}}\cite{guidotti2018survey}: a survey,  \cite{ribeiro2016should}: LIME, \cite{ribeiro2018anchors}: Anchors, \cite{lundberg2017unified}: SHAP, \\ \cite{Kim2018interpretabilityTCAV}: TCAV, \cite{herlocker2000explaining}: explaining recommendation systems, \cite{caruana2015intelligible}: intelligible models, \\ \cite{koh2017understanding}: influence function, \cite{wachter2017counterfactual}: counterfactual explanations, \cite{sundararajan2017axiomatic}: Integrated Gradient (IG),\\ \cite{simonyan2013deep}: saliency maps for images, \cite{selvaraju2017grad}: GradCAM
 \end{tabular} \\\hline
 
 Principles of Explanations &  \begin{tabular}[c]{@{}c@{}}   \cite{kulesza2015principles,kulesza2013too}: completeness and soundness,  \\ \cite{kulesza2012tell}: helping users build mental models \end{tabular} \\ \hline 
 
 User studies for ML & \cite{cai2019human}: image retrieval algorithm for medical uses, \cite{krause2016interacting}: interactive model \\ \hline 
 
 User studies for XAI &  \begin{tabular}[c]{@{}c@{}}  \cite{binns2018s}: justice perceptions, \cite{dodge2019explaining}: fairness \cite{lai2019human}: human-AI team, \cite{hohman2019gamut}: usability, \\ \cite{rader2018explanations,kim2016examples,poursabzi2021manipulating, lim2009and, narayanan2018humans}: understanding, \cite{cheng2019explaining, kaur2020interpreting, buccinca2020proxy,cai2019effects}: trust and understanding  \end{tabular}\\ \hline 
 
 Trust &  \begin{tabular}[c]{@{}c@{}} \cite{bussone2015role}: trust (calibration), \cite{lee2004trust}: trust in automation, \cite{yin2019understanding}: impact of model accuracy on trust, \\
 \cite{kizilcec2016much,cramer2008effects}: impact of system transparency on trust,
 \end{tabular} \\ \hline
\end{tabular}
\caption{Fundamental works of the core papers (categorized according to topics).}
\label{tab:foundation}
\end{table*}
\section{Models and Explanations in XAI User Studies}
\label{sec:modelexplanations}
\begin{figure*}[t]
    \centering
\resizebox{.9\linewidth}{!}{
\begin{tikzpicture}
\usetikzlibrary{calc}

\coordinate (start) at (-4,0);
\coordinate (end) at (18,0);
\draw [line width=2pt, -stealth] (start) -- (end);

\coordinate (s0) at (-3,0);
\coordinate (t0) at ($(s0)+(0,0.3)$);
\coordinate (s1) at (0,0);
\coordinate (t1) at ($(s1)+(0,0.3)$);
\coordinate (s2) at (3,0);
\coordinate (t2) at ($(s2)+(0,0.3)$);
\coordinate (s3) at (6,0);
\coordinate (t3) at ($(s3)+(0,0.3)$);
\coordinate (s4) at (9,0);
\coordinate (t4) at ($(s4)+(0,0.3)$);
\coordinate (s5) at (12,0);
\coordinate (t5) at ($(s5)+(0,0.3)$);
\coordinate (s6) at (15,0);
\coordinate (t6) at ($(s6)+(0,0.3)$);
\coordinate (s7) at (18,0);
\coordinate (t7) at ($(s7)+(0,0.3)$);

\draw [line width=2pt] (s0) -- (t0);
\node [anchor=south] at (t0.north) {$2015$};

\draw [line width=2pt] (t1) -- (s1);
\node [anchor=south] at (t1.north) {$2016$};

\draw [line width=2pt] (t2) -- (s2);
\node [anchor=south] at (t2.north) {$2017$};

\draw [line width=2pt] (t3) -- (s3);
\node [anchor=south] at (t3.north) {$2018$};

\draw [line width=2pt] (t4) -- (s4);
\node [anchor=south] at (t4.north) {$2019$};

\draw [line width=2pt] (t5) -- (s5);
\node [anchor=south] at (t5.north) {$2020$};

\draw [line width=2pt] (t6) -- (s6);
\node [anchor=south] at (t6.north) {$2021$};

\node [anchor=north, align=center, text width=3cm] at (s0.south) {
LRP~\cite{bach2015pixel}
};

\node [anchor=north, align=center, text width=3cm] at (s1.south) {
LIME~\cite{ribeiro2016should} \\
};

\node [anchor=north, align=center, text width=3cm] at (s2.south) {
SHAP~\cite{lundberg2017unified} \\
GradCAM~\cite{selvaraju2017grad} \\
IG~\cite{sundararajan2017axiomatic} \\
SmoothGrad~\cite{smilkov2017smoothgrad}
};

\node [anchor=north, align=center, text width=3cm] at (s3.south) {
INN~\cite{jacobsen2018revnet} \\
TCAV~\cite{Kim2018interpretabilityTCAV}
};

\node [anchor=north, align=center, text width=3cm] at (s4.south) {
ConceptSHAP~\cite{yeh2019completeness}\\
ProtoNet~\cite{chen2019looks}
};

\node [anchor=north, align=center, text width=3cm] at (s5.south) {
MAME~\cite{natesan2020model}\\
Dr.XAI~\cite{panigutti2020doctor}

};
\node [anchor=north, align=center, text width=3cm] at (s6.south) {
SECA~\cite{balayn2021you} \\
VIBI~\cite{bang2021explaining} \\
CLUE~\cite{antoran2021getting}
};

\end{tikzpicture}
}
\caption{\revise{Chronology of commonly used XAI methods from reviewed papers.}}
\label{fig:timetable xai}
\end{figure*}
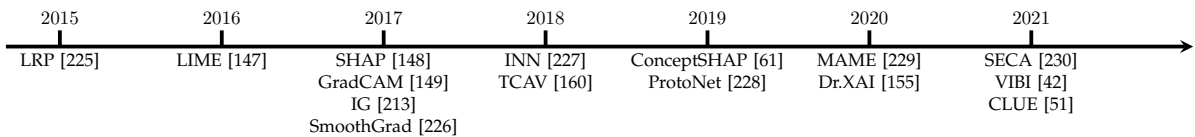

Black-box models are dominant in the current human-AI interaction research area as we can see that more black-box models are studied. Local feature explanations are popularly used such as LIME \cite{ribeiro2016should} and SHAP \cite{lundberg2017unified}. \revise{\Cref{fig:timetable xai} demonstrates the chronological overview of frequently adopted XAI techniques for black-box models in user studies from the surveyed papers}.
However, there are many specific explanation types for certain applications. For recommendation systems, content-based and hybrid explanations are widely used explanations. A content-based explanation is a single-style explanation coming from a content-based recommendation system, while a hybrid explanation contains multiple explanation styles such as user-based or item-based, which is provided by a hybrid recommendation system \cite{kouki2019personalized,friedrich2011taxonomy,kouki2017user}. For instance, 
Dominguez et al.~\cite{dominguez2019effect} provide a content-based explanation as \textit{``Painting A is 85\% similar to the Painting B that you like"}. Tsai et al.~\cite{tsai2019explaining}, however, use hybrid explanations in textual and visual explanation formats.

\section{Measurement Details}

\begin{table}[]
    \centering
     \setlength{\belowcaptionskip}{-5mm}
    \resizebox{\linewidth}{!}{
    \begin{tabular}{ccccc}
    \toprule
       \textbf{Tasks} & \textbf{Tabular} & \textbf{Image/Video} & \textbf{Text} & \textbf{Other} \\
        \midrule
        forward simulation &
        \begin{tabular}[c]{@{}c@{}}
        \cite{hase2020evaluating, poursabzi2021manipulating, ramamurthy2020model} \\
        \cite{antoran2021getting, ribeiro2018anchors, chromik2021think} \\
        \cite{bell2022accuracyexplain, wang2021explanations, cheng2019explaining}
        \end{tabular}
        & \begin{tabular}[c]{@{}c@{}}
        \cite{buccinca2020proxy, nourani2021anchoring,borowski2021exemplary} \\
        \cite{alqaraawi2020evaluating, chandrasekaran2018explanations, colin2022what}\\
        \cite{kim2022hive, shen2020useful}\cite[(VQA)]{ribeiro2018anchors}
        \end{tabular}
        & \cite{hase2020evaluating, arora2022explain} & \begin{tabular}[c]{@{}c@{}} \cite{zhang2022towards} \\ (Audio) \end{tabular}\\ \hline
        \begin{tabular}[c]{@{}c@{}} marginal \\feature effects \end{tabular} & \begin{tabular}[c]{@{}c@{}} \cite{abdul2020cogam, hase2020evaluating, bove2022contextualization} \\ \cite{wang2021explanations, cheng2019explaining} \end{tabular}& & \cite{hase2020evaluating}& \\ \hline
        \begin{tabular}[c]{@{}c@{}} manipulation / \\ counterfactual sim. \end{tabular}& \cite{cheng2019explaining, wang2021explanations, plumb2020regularizing} & \cite{ross2021evaluating} & \cite{arora2022explain} & \\ \hline
        feature importance & \cite{bove2022contextualization, bell2022accuracyexplain, wang2021explanations} & \cite{sixt2022do, Kim2018interpretabilityTCAV}& & \\\hline
        failure prediction & & \cite{chandrasekaran2018explanations}& & \\\hline
        \begin{tabular}[c]{@{}c@{}} relative simulation \\ (selection) \end{tabular} & \cite{chromik2021think, cheng2019explaining} & & & \\\hline
        other & \begin{tabular}[c]{@{}c@{}}\cite{abdul2020cogam} \\ (mental model \\ faithfulness) \end{tabular}& \begin{tabular}[c]{@{}c@{}} \cite{nourani2021anchoring} \\ (class-wise \\ acc.) \end{tabular} & & \\
    \bottomrule
    \end{tabular}
    }
    \caption{Works measuring objective understanding grouped by proxy task/data modality}
    \label{tab:proxytasks}
\end{table}

\begin{table*}[h]
\resizebox{\linewidth}{!}{
\begin{tabular}{c|c|c|c|c}
\specialrule{.1em}{.05em}{.05em}
& \begin{tabular}[c]{@{}c@{}} \textbf{Studied} \\ \textbf{Paper} \end{tabular} & \textbf{Metric} & \begin{tabular}[c]{@{}c@{}} \textbf{Definition} \\ \textbf{Source} \end{tabular} & \textbf{Detail}  \\ \specialrule{.1em}{.05em}{.05em}
\multirow{3}{*}{\textbf{Observed}}   & \cite{panigutti2022understanding} &    Weight of Advice (WOA)  &    -  & \begin{tabular}[c]{@{}c@{}}Degree to which the algorithmic suggestion\\  influences the participant’s estimate.\end{tabular} \\ \cline{2-5} 
 & \cite{suresh2022intuitively, wang2021explanations, schaffer2019can, lai2019human, zhang2020effect, rong2022user}  & Agreement rate   &   - &   \begin{tabular}[c]{@{}c@{}} Percentage of cases in which participants agree with the model. \\ \cite{wang2021explanations} defines the \textit{appropriate trust, overtrust} and \textit{undertrust}.  \\ \cite{schaffer2019can} defines as \textit{adherence} \end{tabular}  \\
 \specialrule{.1em}{.05em}{.05em}
 
\multirow{17}{*}{\textbf{Self-reported}} & \cite{anik2021data, colley2021effects, kaur2020interpreting}  &  \begin{tabular}[c]{@{}c@{}} Trust in \\ Automation \end{tabular} &  \cite{jian2000foundations} & 
\begin{tabular}[c]{@{}c@{}} On the 7-point Likert scale. \\ \cite{anik2021data} adapts the questions. \end{tabular} \\ \cline{2-5} 

& \cite{guo2022building, kunkel2019let} &  \begin{tabular}[c]{@{}c@{}} General trust in technology \end{tabular}  &  \cite{knijnenburg2012explaining}   &  On the 5-point Likert scale. \\ \cline{2-5} 

  & \cite{liao2021should} &  \begin{tabular}[c]{@{}c@{}} Human-Computer \\ Trust \end{tabular}    &    \cite{madsen2000measuring}  & \begin{tabular}[c]{@{}c@{}} On the 7-point Likert scale. \\ \cite{liao2021should} adapts the questions. \end{tabular}    \\ \cline{2-5} 
   
& \cite{ooge2022explaining} &  \begin{tabular}[c]{@{}c@{}} Trust-TAM  (Technology \\ Acceptance Model) \end{tabular}  &  \cite{benbasat2005trust}   &  \begin{tabular}[c]{@{}c@{}} On the 7-point Likert scale. \\ \cite{ooge2022explaining} includes other self-designed questions. \end{tabular} \\  \cline{2-5}

& \cite{cheng2019explaining} &  \begin{tabular}[c]{@{}c@{}} Trust in human-machine systems \end{tabular}  &  \cite{lee1992trust}   &  \begin{tabular}[c]{@{}c@{}} On the 7-point Likert scale. \end{tabular} \\  \cline{2-5}  

& \cite{ehsan2019automated} & \begin{tabular}[c]{@{}c@{}} Unified Theory of Acceptance \\ and Use of Technology Model (UTAUT) \end{tabular} &  \cite{venkatesh2003user} & On the 5-point Likert scale. \\ \cline{2-5} 

& \cite{paleja2021utility} & \begin{tabular}[c]{@{}c@{}} Human-Robot Collaborative\\ Fluency Assessment \end{tabular} &  \cite{hoffman2019evaluating} & On the 7-point Likert scale \\ \cline{2-5} 

& \cite{kunkel2019let}  & \begin{tabular}[c]{@{}c@{}} Trusting beliefs\\ and intentions \end{tabular}    & \cite{mcknight2002developing}                                                         &  On the 7-point Likert scale. \\  \cline{2-5}  

 &   \begin{tabular}[c]{@{}c@{}} \cite{tsai2021exploring, buccinca2020proxy, schoeffer2022there, schaffer2019can, smith2020digging, smith2020no} \\ \cite{dominguez2019effect, cai2019effects, millecamp2019explain, tsai2018beyond, kim2020answering, rong2022user} \end{tabular} &  \begin{tabular}[c]{@{}c@{}}  Self-designed \\ questionnaire \end{tabular}   & -   &  \begin{tabular}[c]{@{}c@{}} \cite{tsai2021exploring, schaffer2019can, cai2019effects} are on the 7-point Likert scale. \\ \cite{buccinca2020proxy, schoeffer2022there, millecamp2019explain, tsai2018beyond,kim2020answering} are on the 5-point Likert scale. \\ \cite{dominguez2019effect} rates from 0 to 100.\\
 \cite{dominguez2019effect, buccinca2020proxy, millecamp2019explain, tsai2018beyond, kim2020answering,smith2020digging,smith2020no} measure one-dimensional trust. \end{tabular} \\ \cline{2-5}

 & \cite{ehsan2021expanding}  & \begin{tabular}[c]{@{}c@{}} Semi-structured \\ interview \end{tabular}  & -  & \\
 \specialrule{.1em}{.05em}{.05em}
\end{tabular}
}
\caption{Measures of trust. The measurement is divided into two main groups: ``Observed'' and ``self-reported'' trust. The studied core papers using the same measurement are grouped together. The name and the paper reference of the used metrics are listed in the column "Metric" and "Definition Source", respectively. ``-'' in the column ``Definition Source'' means that the source is the studied paper.  More details about the metrics are given in the last column.}
\label{tab:trust measures}
\end{table*}

\subsection{Trust}
\Cref{tab:trust measures} lists the trust measurement. Most of the works deploy questionnaires to measure user trust (self-reported), where a 7-point or 5-point Likert scale is commonly used. Many works design their own questionnaires \cite{tsai2021exploring, buccinca2020proxy, schoeffer2022there, schaffer2019can,dominguez2019effect, cai2019effects, millecamp2019explain, tsai2018beyond, kim2020answering}. To measure trust in an objective manner, many works choose to use the agreement rate of humans \cite{suresh2022intuitively, wang2021explanations, schaffer2019can, lai2019human}.

\subsection{Usability}
\Cref{tab:usability measures} demonstrates the measures used for the usability of explanations. We divide usability into five sub-categories: workload (cognitive load), helpfulness, satisfaction, \revise{undesired behavior detection} and ease of use and others. User perceptions of workload, helpfulness, satisfaction and ease of use are subjective and often measured with questionnaires. However, for debugging tasks, it can be measured objectively such as using the accuracy of the user confirming the correctness of answers from a question-answering model and the time for solving this task \cite{kim2020answering}.

\begin{table*}[h]
\resizebox{\linewidth}{!}{
\begin{tabular}{c|c|c|c|c}
\specialrule{.1em}{.05em}{.05em}
& \begin{tabular}[c]{@{}c@{}} \textbf{Studied} \\ \textbf{Paper} \end{tabular}
& \textbf{Metric} 
& \begin{tabular}[c]{@{}c@{}} \textbf{Definition} \\ \textbf{Source} \end{tabular} 
& \textbf{Detail}  \\ \specialrule{.1em}{.05em}{.05em}

\multirow{2}{*}{\begin{tabular}[c]{@{}c@{}}\textbf{Workload}\end{tabular}}
&  \cite{kaur2020interpreting, dominguez2019effect, colley2021effects, arendt2020parallel, springer2019progressive}
& NASA TLX
& \cite{hart1988development}
&  \\ \cline{2-5}

&  \cite{abdul2020cogam}
& Memory Performance
&     -
&  \\ \specialrule{.1em}{.05em}{.05em}

\multirow{2}{*}{\textbf{Helpfulness}}
& \begin{tabular}[c]{@{}c@{}} \cite{buccinca2020proxy, nourani2021anchoring, wang2022interpretableideation} \\ \cite{zhang2022debiased, zhang2022towards, abdul2020cogam} \end{tabular}
& \begin{tabular}[c]{@{}c@{}} Self-designed \\ questionnaire \end{tabular}
&  -
& \begin{tabular}[c]{@{}c@{}} \cite{buccinca2020proxy, nourani2021anchoring, wang2022interpretableideation} are on 5-point Likert scale \\
\cite{abdul2020cogam, zhang2022debiased, zhang2022towards} are on 7-point Likert scale \end{tabular}
\\ \cline{2-5}

& \cite{gao2019explainable}
&  Rating
&  -
&  Rating from 1 to 5
\\ \cline{2-5}

& \cite{plumb2020regularizing}
&  \revise{Comparison}
&  -
&  \revise{Users select the most helpful method}
\\ \specialrule{.1em}{.05em}{.05em}

\multirow{7}{*}{\textbf{Satisfaction}}
& \cite{dominguez2019effect, millecamp2019explain, smith2020digging, smith2020no,tsai2018beyond,tsai2021exploring,tsai2019explaining}
& \begin{tabular}[c]{@{}c@{}} Self-designed \\ questionnaire \end{tabular}
& -
& \begin{tabular}[c]{@{}c@{}} \cite{millecamp2019explain,tsai2018beyond,tsai2019explaining} are on 5-point Likert scale \\
\cite{smith2020no, smith2020digging,tsai2021exploring} are on 7-point Likert scale \\
\cite{dominguez2019effect} rates from 0 to 100
\end{tabular}   
\\ \cline{2-5}

& \cite{guo2022building,kouki2019personalized} & \begin{tabular}[c]{@{}c@{}} User experience of \\ recommendation system  \end{tabular}& \cite{knijnenburg2012explaining} &  \begin{tabular}[c]{@{}c@{}} \cite{guo2022building} adapts the questions on the 5-point Likert scale \\  \cite{kouki2019personalized} adapts the questions on the 7-point Likert scale \end{tabular} \\\cline{2-5}


& \cite{bove2022contextualization, panigutti2022understanding}
& \begin{tabular}[c]{@{}c@{}} Explanation \\ Satisfaction Scale  \end{tabular}
& \cite{hoffman2018metrics}
&  \begin{tabular}[c]{@{}c@{}} \cite{panigutti2022understanding} are on 5-point Likert scale \\ 
\cite{bove2022contextualization} are on 6-point Likert scale
\end{tabular}
\\ \specialrule{.1em}{.05em}{.05em}

\multirow{12}{*}{\begin{tabular}[c]{@{}c@{}}\textbf{Undesired} \\ \textbf{behavior} \\\textbf{detection}\end{tabular}}
& \cite{balayn2022can}
& \begin{tabular}[c]{@{}c@{}} Number of \\ identified bugs \end{tabular}
& -
& Questions about bug identification and solutions
\\ \cline{2-5}
& \cite{kim2020answering}
&  \begin{tabular}[c]{@{}c@{}} Accuracy (percentage of \\ correct answers) and time \end{tabular}
& -
& \begin{tabular}[c]{@{}c@{}} Task is to determine the \\correctness of model answers \end{tabular}
\\ \cline{2-5}
& \cite{poursabzi2021manipulating}
& \begin{tabular}[c]{@{}c@{}}Deviation between human's \\ and model's predictions \end{tabular}
& -
&\begin{tabular}[c]{@{}c@{}} Model's predictions are buggy and \\ human's predictions should be different.\end{tabular}
\\ \cline{2-5}
& \cite{sixt2022do} & \begin{tabular}[c]{@{}c@{}} Accuracy (percentage of\\ correct answers)  \end{tabular} & - &  \begin{tabular}[c]{@{}c@{}} Task is to identify \\ (ir)relevant features \end{tabular}
\\ \cline{2-5}
& \cite{rawal2020beyond} & \begin{tabular}[c]{@{}c@{}}   Accuracy of \\ answers \end{tabular} & - &  \begin{tabular}[c]{@{}c@{}} Task is to  detect model biases \\
or discrimination \end{tabular} \\ \cline{2-5}

& \cite{schoeffer2022there,rader2018explanations, anik2021data, dodge2019explaining,binns2018s,schoeffer2021appropriate} & \begin{tabular}[c]{@{}c@{}}   Rating \end{tabular} & - &  \begin{tabular}[c]{@{}c@{}} \cite{schoeffer2022there}: to judge whether they receive enough information  \\ to judge the model process is unfair or not; \\ The other judge the model is unfair or not. \end{tabular} \\ \cline{2-5}

& \cite{grgic2018human} & \begin{tabular}[c]{@{}c@{}} Rating \end{tabular} & - &  \begin{tabular}[c]{@{}c@{}}  Rating on the unfairness of features\ \end{tabular} \\ \cline{2-5}
\specialrule{.1em}{.05em}{.05em}

\multirow{14}{*}{\begin{tabular}[c]{@{}c@{}} \textbf{Ease of use} \\ \textbf{and others} \end{tabular}}
& \cite{abdul2020cogam, hohman2019gamut, wang2022interpretableideation, kim2020answering, arendt2020parallel}
& \begin{tabular}[c]{@{}c@{}} Self-designed \\ questionnaire \end{tabular}
&  -
& \begin{tabular}[c]{@{}c@{}} \cite{wang2022interpretableideation, kim2020answering} are on 5-point Likert scale \\
\cite{abdul2020cogam, hohman2019gamut} are on 7-point Likert scale
\end{tabular}
\\ \cline{2-5}

& \cite{schneider2021explain}
& \begin{tabular}[c]{@{}c@{}} AVAM and UEQ-S \end{tabular}
&  \cite{hewitt2019assessing, schrepp2017design}
&  \begin{tabular}[c]{@{}c@{}} Autonomous Vehicle Acceptance Model Questionnaire (AVAM) \cite{hewitt2019assessing} \\ 
User Experience Questionnaire-Short (UEQ-S) \cite{schrepp2017design} \\
Both on the 7-point Likert scale
\end{tabular}
\\ \cline{2-5}

& \cite{ross2021evaluating}
& \begin{tabular}[c]{@{}c@{}} Single Ease \\ Question (SEQ) \end{tabular}
& \cite{sauro2009comparison}
&  On the 7-point Likert scale \\ \cline{2-5}

& \cite{balayn2022can}
& \begin{tabular}[c]{@{}c@{}} User Engagement \\ Scale (UES) \end{tabular}
& \cite{o2018practical}
&  On the 7-point Likert scale \\ \cline{2-5}

& \cite{kuhl2022keep, kuhl2022let}
& \begin{tabular}[c]{@{}c@{}} System Causability \\ Scale \end{tabular}
& \cite{holzinger2020measuring}
&  On the 5-point Likert scale \\ \cline{2-5}

& \cite{colley2021effects}
& \begin{tabular}[c]{@{}c@{}} System Usability \\ Scale \end{tabular}
& \cite{brooke1996sus}
&  On the 5-point Likert scale \\\cline{2-5}

& \cite{le2018improving,li2019data}
& \begin{tabular}[c]{@{}c@{}} semi-structured interview \end{tabular}
&  -
&  -
\\
\specialrule{.1em}{.05em}{.05em}

\end{tabular}
}
\caption{Measures of usability. The measurement is divided into five categories. The studied core papers using the same measurement are grouped together. The name and the paper reference of the used metrics are listed in the column "Metric" and "Definition Source", respectively. ``-'' in the column ``Definition Source'' means that the source is the studied paper.  More details about the metrics are given in the last column.}
\label{tab:usability measures}
\end{table*}

\subsection{Understanding of Explanations}
\label{sec:understandingofexplanations}
For novel or cognitively challenging types of explanations, it makes sense to verify whether users can make use of the information provided through the explanation. Usually these types of tests are conducted in combination with other measures to establish if the explanations are correctly understood by users and can thus be processed as intended.

In the domain of conceptual explanations \cite{Kim2018interpretabilityTCAV, koh2020concept}, such kind of understanding questions are common, to assess semantic coherence of automatically discovered concepts \cite{laina2020quantifying, yeh2019completeness, ghorbani2019towards, leemann2022coherence}. Assignment tasks, where novel instances should be assigned to existing clusters are commonly used as a proxy to measure the intelligibility \cite{laina2020quantifying, ramamurthy2020model, yeh2019completeness, ghorbani2019towards}. Another option is to assess how well the cluster can be described in natural language which is often referred to as \textit{describability} \cite{ghorbani2019towards, laina2020quantifying, leemann2022coherence}.
Apart from conceptual explanations, Zhang et al.~\cite{zhang2022towards} ask multiple choice questions to verify if users understand the differences between the acoustical cues presented and evaluate which cue differences were most noticeable. Wang et al.~\cite{wang2022interpretableideation} prompt users explicitly if the found the explanation easy to understand. 

\paragraph*{Research questions and Findings} Laina et al.~\cite{laina2020quantifying} found that feature vectors obtained by contrastive learning approaches such as MoCo \cite{he2020momentum} or SeLa \cite{asano2020Self-labelling} allow for clusters that are almost as interpretable as human labels. Leemann et al.~\cite{leemann2022coherence} show the similarity of ResNet-50 embeddings allows to predict how semantically coherent users find a cluster of images.
For the acoustical cue, Zhang et al.~\cite{zhang2022towards} found that shrillness and speaking rate were most often recognized. Wang et al.~\cite{wang2021explanations} found that users reported they understood all types of explanations well without significant differences.

\section{Findings}
\label{sec:appfindings}
When using explanation types as the evaluation dimension, many works compare their effects without comparing them to a control group (baseline) without explanation methods. Anik et al.~\cite{anik2021data} argue that many works have proven the usefulness of explanations and therefore no need to include such a control group. \Cref{tab:findings supp} summarizes the findings of the comparison among different explanations. \Cref{tab:findings2} lists results of using other evaluation dimensions beyond explanations.

\begin{table*}[h]
\centering
 \setlength{\belowcaptionskip}{-5mm}
\resizebox{.9\linewidth}{!}{
\begin{tabular}{cc|cc}
\specialrule{.1em}{.05em}{.05em}

\multicolumn{2}{c|}{\multirow{2}{*}{}}&  \multicolumn{2}{c}{\multirow{1}{*}{\textbf{Other Evaluation Dimensions}}}                         \\ \cline{3-4}
\multicolumn{2}{c|}{}   
& \multicolumn{1}{c|}{Positive} & \multicolumn{1}{c}{Non-positive / Mixed}
\\ \specialrule{.1em}{.05em}{.05em}

\multicolumn{2}{c|}{\textbf{Trust}}  & \multicolumn{1}{c|}{\begin{tabular}[c]{@{}c@{}} \cite{anik2021data}: balanced training data, \\\cite{cai2019effects}: high model performance \\  \cite{kunkel2019let}: high quality of explanations \\
 \cite{schoeffer2022there}: high AI literacy \\
 \cite{smith2020no}: interactivity \\
 \revise{\cite{zhang2020effect}: model confidence} \\
 \end{tabular}}  
 & \multicolumn{1}{c}{\begin{tabular}[c]{@{}c@{}}\cite{anik2021data}: user expertise, insignificant \\ \cite{millecamp2019explain}: personal characteristics, insignificant \\ \cite{smith2020digging}: different topic modeling approaches, insignificant \\
 \underline{\cite{liao2021should}: self-referential pronoun ``I'' in explanations, negative} \\
 \cite{cheng2019explaining}: user technical literacy, insignificant \\
\end{tabular}}  \\ \specialrule{.1em}{.05em}{.05em}
\multicolumn{1}{c|}{\multirow{2}{*}{\textbf{Understanding}}}               & Obj.    & \multicolumn{1}{c|}{\begin{tabular}[c]{@{}c@{}}
\cite{ross2021evaluating}: disentanglement  of gen. model \\
\cite{cheng2019explaining}: interactivity \\
\revise{\cite{plumb2020regularizing}: ExpO regularization of the model} \end{tabular}}
& \multicolumn{1}{c}{\begin{tabular}[c]{@{}c@{}} 
\underline{\cite{ross2021evaluating, poursabzi2021manipulating}: high dimensionality, negative}\\
\cite{bove2022contextualization}: contextualization, insignificant \\
 \cite{buccinca2020proxy}: inductive vs. deductive explanations, insignificant \\
\cite{arora2022explain}: different ML models, insignificant \\
\cite{borowski2021exemplary}: user expertise, insignificant \\
\cite{chandrasekaran2018explanations}: instant feedback, insignificant \\
 \cite{nourani2021anchoring}: timing of model errors, mixed\\

\end{tabular}
}   \\ \cline{2-4} 
\multicolumn{1}{c|}{}  & Sub. & 
\multicolumn{1}{c|}{\begin{tabular}[c]{@{}c@{}} 
  \cite{ross2021evaluating}: disentanglement of gen. model \\
  \cite{cheng2019explaining}: interactivity \\
  \revise{\cite{szymanski2021visual}: user expertise}
\end{tabular}}  
& \multicolumn{1}{c}{\begin{tabular}[c]{@{}c@{}} 
    \cite{hase2020evaluating}: model correctness, insignificant \\
    \cite{rebanal2021xalgo}: QuickSort, insignificant \\
    \cite{chromik2021think}: \underline{test of understanding, negative} \\

\end{tabular}}   \\ \specialrule{.1em}{.05em}{.05em}

\multicolumn{2}{c|}{\textbf{Usability}}  & 
\multicolumn{1}{c|}{\begin{tabular}[c]{@{}c@{}} 
    \cite{ross2021evaluating}: significant difference in self-reported \\ difficulty dependent on the generative model \\
    \cite{guo2022building, smith2020no}: interactivity \\
    \cite{arendt2020parallel}: Parallel Embeddings \\
    \cite{schoeffer2022there}: high AI literacy \\
    \cite{grgic2018human}: fair features are ``current charges'' \\ and ``criminal history''
\end{tabular}} 
& \multicolumn{1}{c}{\begin{tabular}[c]{@{}c@{}} 
    \cite{smith2020digging}: different topic modeling approaches, insignificant \\
    \cite{millecamp2019explain}: personal characteristics, insignificant for satisfaction \\
    \underline{\cite{nourani2021anchoring}: early encounters of system weaknesses}
    \\\underline{lead to lower explanation usage} \\
    \cite{poursabzi2021manipulating}: clear model is \textit{less} useful in debugging \\
     \cite{grgic2018human}: \underline{unfair} features are ``quality of school life'' \\and ``education \& school behavior'', etc.
\end{tabular}} \\ 
\specialrule{.1em}{.05em}{.05em}
\multicolumn{2}{c|}{\begin{tabular}[c]{@{}c@{}}\revise{\textbf{Human-AI}}\\ 
\revise{\textbf{Collaboration Performance}}\end{tabular}}
& \multicolumn{1}{c|}{\begin{tabular}[c]{@{}c@{}}\cite{lai2020chicago}: low model complexity \\
\cite{alufaisan2021does, buccinca2020proxy}: showing model prediction 
\end{tabular}} 
& \multicolumn{1}{c}{\begin{tabular}[c]{@{}c@{}} 
 \cite{paleja2021utility}: explanations are positive for novices' performance \\but \underline{negative for experts'}\\
 \revise{\cite{kim2022hive, zhang2020effect} Showing predictions, insignificant}
 \revise{\cite{zhang2020effect}: model confidence}
\end{tabular}}  \\ \specialrule{.1em}{.05em}{.05em}
\end{tabular}
}
\caption{User study findings when using \textbf{other aspects} (other than the presence of explanation) as evaluation dimensions. Effects on measured quantities are divided into ``Positive'' where explanation information is given, and ``Non-positive / Mixed'' where negative impact is marked with \underline{underlines}.}
\label{tab:findings2}
\end{table*}

\begin{table*}[]
\resizebox{.9\linewidth}{!}{
\begin{tabular}{cc|c}
\specialrule{.1em}{.05em}{.05em}
\multicolumn{2}{c|}{\multirow{2}{*}{}} & \multicolumn{1}{c}{\textbf{Evaluation Dimension: Explanations}}                        \\ 
\multicolumn{2}{c|}{\multirow{2}{*}{}}   & \multicolumn{1}{c}{Effect comparison among \textbf{different explanations}}
\\\specialrule{.1em}{.05em}{.05em}
\multicolumn{2}{c|}{Trust}  
&  \begin{tabular}[c]{@{}c@{}} 
\cite{suresh2022intuitively}: example-based explanations are positive in trust building \\
\cite{buccinca2020proxy}: deductive (rule-based) explanations $>$ inductive (example-based) explanations \\in decision-making tasks, 
but contrary in proxy tasks \\
\cite{cai2019effects}: different explanations positively affect different beliefs of trust \\
\cite{tsai2018beyond}: proposed explanation interfaces (different visualizations), \\ SCATTER $>$ RANK and SCATTER $>$ TUNER but insignificant\\
\revise{\cite{peng2022inherently} HEX-RL (theirs) > LSTM-attention (for RL agents)}
\end{tabular}  \\ \specialrule{.1em}{.05em}{.05em}
\multicolumn{1}{c|}{\multirow{2}{*}{Understanding}}                                      & Obj. 
&
\begin{tabular}[c]{@{}c@{}}\cite{zhang2022towards}: Cues and Counterfactuals > Saliency (audio data) \\ \cite{abdul2020cogam}: Sparse Lin. > COGAM > GAM \\ \cite{ramamurthy2020model}: MAME  > SP-LIME\\
\cite{antoran2021getting}: CLUE > Sensitivity, Human CLUE, Random (for uncertainty)\\
\cite{borowski2021exemplary}: Natural images > synthetic (activation prediction)\\
\cite{sixt2022do}: Counterfactuals (INN) = (proposed) Baseline Expl. > Concepts\\
\cite{ribeiro2018anchors} Anchors > LIME\\
\end{tabular}  \\ \cline{2-3} 
\multicolumn{1}{c|}{}    
& Sub. 
& \multicolumn{1}{c}{\begin{tabular}[c]{@{}c@{}} 
\cite{radensky2022exploring}: local+global explanation > local/global explanation \\
\cite{cai2019effects}: example-based explanations (normative/comparative) 
improve the subj. understanding \\
\cite{hase2020evaluating} LIME $\geq$ Composite, Prototypes and others \\
\cite{kuhl2022keep}: closest and plausible counterfactuals, difference insignificant \\
\cite{radensky2022exploring}: local+global explanation > local/global explanation \\
\revise{\cite{peng2022inherently} HEX-RL (theirs) > LSTM-attention (for RL agents)}  \\
\revise{\cite{szymanski2021visual}: visual > textual explanations} 
\end{tabular}} \\
\specialrule{.1em}{.05em}{.05em}
\multicolumn{2}{c|}{Usability}        
& 
\multicolumn{1}{c}{\begin{tabular}[c]{@{}c@{}} 
    \cite{abdul2020cogam}: sLM $\leq$ COGAM $<$ GAM, 
    insignificant for self-reported cognitive load \\
    \cite{bove2022contextualization}: contextualizing/exploration improve user's satisfaction, \\
    but no significant impact when interacting both factors \\
    \cite{hohman2019gamut}: diff. expl. (e.g. local expl., counterfactuals,...) \\
    \cite{kaur2020interpreting}: GAM vs. SHAP, pos. for cognitive load \\
    \cite{dominguez2019effect}: diff. interfaces, pos. for cognitive load \\
    \cite{zhang2022towards}: counterfactual+cues > saliency, pos. for helpfulness\\
    \cite{gao2019explainable}: DEAML $>$ EFM (feature-level expl.) $>$ PAV (``people also viewed'' expl.) for usefulness in RS \\
    \cite{nourani2021anchoring}: Salient video segments $>$ Confidence scores, \\ Component combinations shown for helpfulness \\
    \cite{buccinca2020proxy}: deductive (rule-based) has higher cognitive load than inductive (example-based) in proxy tasks, \\
   deductive (rule-based) $>$ inductive (example-based) in helpfulness in decision-making task \\
    \cite{kuhl2022keep}: closest and plausible counterfactuals, difference insignificant \\
    \cite{kouki2019personalized}: text explanation $>$ visual explanations in user experience (e.g., satisfaction) \\
    \cite{tsai2018beyond}: proposed explanation interfaces (different visualizations), \\ SCATTER $>$ RANK and TUNER $>$ SCATTER in satisfaction, \\ RANK $>$ SCATTER and TUNER $>$ SCATTER in usefulness, but all insignificant \\
    \cite{sixt2022do}: Counterfactuals (INN) = (proposed) Baseline Expl. > Concepts in bias detection \\
    \revise{\cite{rawal2020beyond}: AReS (theirs) > AR-LIME}\\
\cite{dodge2019explaining}: sensitivity- and case-based explanations are rated as least fair when they expose a bias of the model\\
\cite{wang2022humans}: acceptance of the gender-aware career recommender > gender-debiased\\
\cite{harrison2020empirical}: significant preference for equalizing false positives over equalizing accuracy\\
\cite{schoeffer2022there}: the amount of information positively relates with perceived fairness\\
\cite{anik2021data}: data-centric explanations that indicate balanced training data raise the fairness rating 
\end{tabular}}  \\
\specialrule{.1em}{.05em}{.05em}
\multicolumn{2}{c|}{\begin{tabular}[c]{@{}c@{}}\revise{Human-AI}\\ \revise{Collaboration Performance}\end{tabular}}    
& \multicolumn{1}{c}{\begin{tabular}[c]{@{}c@{}} \cite{buccinca2020proxy}: both deductive (rule-based) explanations and inductive (example-based) explanations \\ are positive, no significant difference \end{tabular}}
\\ \specialrule{.1em}{.05em}{.05em}
\end{tabular}}
\caption{User study findings when using model \textbf{explanations} as evaluation dimensions and comparing different explanation types on measured quantities.}
\label{tab:findings supp}
\end{table*}

\section{Towards Increasingly User-Centered XAI}
\label{sec:towardshxai}
In this section, we provide a detailed literature review regarding existing work in pedagogical
frameworks, which provides implications for designing future transparent AI
systems and human-centered evaluations in Sec. 7.1.
\subsection{Expectancy-value Motivation Theory}
\label{sec:expectancyvaluetheory}
\added{
Human interaction with XAI interfaces can be viewed as an activity where humans learn about the model's inner workings through explanations and then achieve an understanding of the models. 
The question of how to enhance the efficiency and the outcome of this human learning process is of high importance \cite{lage2019exploring}. This research question is widely considered in educational psychology through the lens of expectancy-value motivation theory \cite{hulleman2017making,richardson2012psychological,wigfield2010expectancy}.
For instance, Hulleman et al.~\cite{hulleman2017making} propose to utilize \textit{interventions} to increase the perception of usefulness (utility value) to subsequently increase motivation and final performance. Intervention here refers to identifying the relevance of model explanations to the user's own situation, which can be a prompt question while working with the interface. Moreover, when utilizing model explanations in human-AI collaboration, explanations can be seen as a type of ``scaffolding'' (prompt during a task) proposed in a conceptual framework in education \cite{chernikova2022theoretical,heitzman2019facilitating}. 
Bisra et al.~\cite{bisra2018inducing} summarize guidelines for effective scaffolding. For instance, different disciplinary descriptions can be used in the scaffolding (explanation prompt) to enhance the user's intuition. Another important, yet often unconsidered point is the role of personality traits in the perception of explanations. For instance, Conati et al.~\cite{conati2021toward} show that the \textit{need for cognition} characteristic, which indicates users' openness towards cognitively challenging tasks, is a determining factor for explanation effectiveness in an intelligent tutoring system. Considering these findings, we see personalized XAI as a \vu{relatively underexplored} but yet sorely needed research direction.}

\vspace{-6pt}
\subsection{Theory of Mind}
\label{sec:theoryofmind}
When interacting with XAI systems, humans form mental models of the machine learning algorithms that reflect their belief of how the algorithms work. The formation of these mental models comes from observing explanations or examples given to the human, who often subconsciously applies the observations in a few examples to the broader understanding of the whole machine learning system. This incredible ability to infer, rationalize, and summarize other intelligent agents' decisions is known as the Theory of Mind (ToM) \cite{becker1976economic, cohen1991precursors} in psychology. 
Based on this theory, Bayesian Theory of Mind (BToM) \cite{baker2012tbayesian} provides a probabilistic framework to predict the inferences that people make about the mental states underlying other agents' actions \cite{csibra2017cognitive}.
Recent work, at the intersection of XAI and robotics, indicates that humans also attribute ToM to artificial agents that they observe or interact with \cite{hellstrom2018understandable, lee2005human}. Guided by these user-centered results, several works at the intersection of XAI and robotics have utilized BToM to create a simulated user and then use the simulated user to generate helpful explanations.
Towards this goal, Huang et al.~\cite{huang2019enabling} provide a greedy algorithm for selecting explanations that maximize the simulated user's knowledge of the agent's (a self-driving car in their domain) policy; and Lee et al.~\cite{lee2021humanIRL} provide a related approach where the user is modeled as an inverse reinforcement learner. In addition to selecting the most informative explanations, Qian and Unhelkar~\cite{qian2022evaluating} utilize a variation of the Monte Carlo tree search to generate a computationally tractable approach to identify the most informative sequence of the explanations, based on the assumption that some explanations might be more effective initially.
Thus, while some existing works evaluate the effectiveness of the selected explanations through experiments with human users, the community still lacks an understanding of how robust or realistic BToM is compared to a human's cognitive process particularly for XAI. We also advocate for more probabilistic and computational cognitive models to be utilized in XAI designs. To achieve this, we need experts from cross disciplines to address individual user's needs in an XAI system from cognitive, psychological, and computational perspectives.
Lastly, we also encourage XAI researchers to develop solutions to explain \textit{AI-enabled systems} -- for instance, robots and autonomous vehicles -- which require grounded and user-centered solutions.

\subsection{Hybrid Teaching}
\label{sec:hybridteaching}
Teaching strategies for the human-to-human setting have been widely studied and many categorizations exist \cite{julienSchultz2010direct, nguyen2017students, kipper2011teaching}. One way of categorizing these strategies is through the following three concepts: (1) direct teaching, (2) indirect teaching, and (3) hybrid teaching.
\textit{Direct teaching} utilizes direct instructions that are teacher-centered, involve clear teaching objectives, and are consistent with classroom organizations. In XAI applications, direct teaching methods generate explanations by selecting representative examples of an agent's decisions to convey the patterns in its policy \cite{amir2018highlights, huang2018establishing, amir2019summarizing, lee2021humanIRL, watkins2021explaining, amitai2022summarizing}.
In contrast, \textit{indirect teaching} is student-centered and encourages independent learning. In the XAI perspective, methods utilizing indirect teaching provide users with tools to actively and independently explore an AI system.
Although the goal of direct and indirect teaching methods is the same, namely explaining an AI system to human users, the computational problems solved by these methods are different.
Direct teaching focuses on providing guidance (using a computational approach) to assist users in building an understanding of a machine, whereas indirect teaching (often through a user interface) enables users to address individual learning preferences and mitigate individual confusion about the AI.
To leverage the advantages of the two teaching strategies, \textit{hybrid teaching} has been widely used in human-to-human teaching with an emphasis on interactivity \cite{fulford1993perceptions, muirhead2001interactivity, muirhead2004interactivity}.
In XAI-related work, Qian and Unhelkar~\cite{qian2022evaluating} provide a hybrid teaching framework by introducing an \textit{AI Teacher} to enable guided interactivity between RL-based AI agents and a user. Their results indicate that hybrid teaching reduces the amount of time for a user to understand an agent's policy compared to direct and indirect teaching, and is more subjectively preferred by the participants.  
Building on this, future XAI systems can consider using hybrid teaching methods that $(i)$ generate direct instructions to provide guidance to users' understanding of an AI system and $(ii)$ provide methods to allow users to interact with the agent or model enabling active learning. 

\end{document}